\documentclass[journal]{IEEEtran}

\PassOptionsToPackage{numbers, compress}{natbib}

\usepackage[utf8]{inputenc} 
\usepackage[T1]{fontenc}    
\usepackage{hyperref}       
\usepackage{url}            
\usepackage{booktabs}       
\usepackage{amsfonts}       
\usepackage{nicefrac}       
\usepackage{microtype}      
\usepackage{xcolor}         
\usepackage{graphicx}
\usepackage{multirow}
\usepackage{amsmath}
\usepackage{colortbl}
\usepackage{xcolor}
\usepackage{floatrow}
\usepackage{wrapfig} 
\usepackage{enumitem}
\usepackage{comment}
\newfloatcommand{capbtabbox}{table}[][\FBwidth]
\usepackage{subcaption}
\usepackage{tabularx}

\newcommand{\mj}[1]{\textcolor{black}{#1}}

\begin{document}

\title{A Benchmark for Cycling Close Pass Detection from Video Streams}

\author{Mingjie Li, Ben Beck, Tharindu Rathnayake,  Lingheng Meng, \\ Zijue Chen, Akansel Cosgun, Xiaojun Chang, Dana Kuli\'c
\thanks{This project was funded by an Australian Government Department of Infrastructure, Transport, Regional Development and Communications `Road Safety Innovation Fund' grant. Ben Beck was supported by an Australian Research Council Future Fellowship (FT210100183). Dana Kuli\'c was supported by an Australian Research Council Future Fellowship (FT200100761). Xiaojun Chang was supported by the Australian Research Council (ARC) Discovery Early Career Researcher Award (DECRA) under DE190100626.} \thanks{Mingjie is with Stanford University, United States. This work was done while Mingjie was with Monash University.} \thanks{Tharindu, Zijue, Lingheng, Ben and Dana are with Monash University, Australia. Akansel Cosgun is with Deakin University, Australia. Xiaojun is with the University of Technology Sydney.}}


\maketitle

\begin{abstract}
Cycling is a healthy and sustainable mode of transport. However, interactions with motor vehicles remain a key barrier to increased cycling participation. The ability to detect potentially dangerous interactions from on-bike sensing could provide important information to riders and policymakers. A key influence on rider comfort and safety is close passes, i.e., when a vehicle narrowly passes a cyclist. In this paper, we introduce a novel benchmark, called Cyc-CP, towards close pass (CP) event detection from video streams. The task is formulated into two problem categories: scene-level and instance-level. Scene-level detection ascertains the presence of a CP event within the provided video clip. Instance-level detection identifies the specific vehicle within the scene that precipitates a CP event. To address these challenges, we introduce \mj{four} benchmark models, each underpinned by advanced deep-learning methodologies. For training and evaluating those models, we have developed a synthetic dataset alongside the acquisition of a real-world dataset. The benchmark evaluations reveal that the models achieve an accuracy of 88.13\% for scene-level detection and 84.60\% for instance-level detection on the real-world dataset. We envision this benchmark as a test-bed to accelerate CP detection and facilitate interaction between the fields of road safety, intelligent transportation systems and artificial intelligence. Both the benchmark datasets and detection models will be available at \url{https://github.com/SustainableMobility/cyc-cp} to facilitate experimental reproducibility and encourage more in-depth research in the field.
\end{abstract}

\begin{IEEEkeywords}
Cycling Safety, Close Pass, Video Action Recognition, Monocular 3D Object Detection
\end{IEEEkeywords}

\section{Introduction}

Cycling is an active mode of transport that confers substantial health, environmental, and social benefits~\cite{celis2017association,mason2015global,cai2023bicycle,li2023improving}. Despite the many benefits, cycling represents only a small proportion of trips in many cities around the world~\cite{goel2022cycling,beck2021spatial}. The major barrier to increased cycling participation is how unsafe people feel when riding~\cite{pearson2022adults}; specifically, fears related to riding in close proximity to motor vehicles, risk of injury, and a lack of appropriate bicycling infrastructure. Cities have often struggled to implement cycling infrastructure that provides physical separation between cyclists and motor vehicles~\cite{wilson2020implementing}, and as such, the majority of on-road infrastructure consists of painted bike lanes (for example, 99\% of on-road infrastructure in Melbourne, Australia~\cite{beck2021spatial}). As a result, interactions between cyclists and motor vehicles are frequent. In a study of lateral passing distance between cyclists and motor vehicles, `dangerous' passing events were frequent and were noted to make cyclists feel unsafe~\cite{beck2019much,beck2021subjective}. 

\begin{figure*}[t]
\centering
\includegraphics[width=\textwidth]{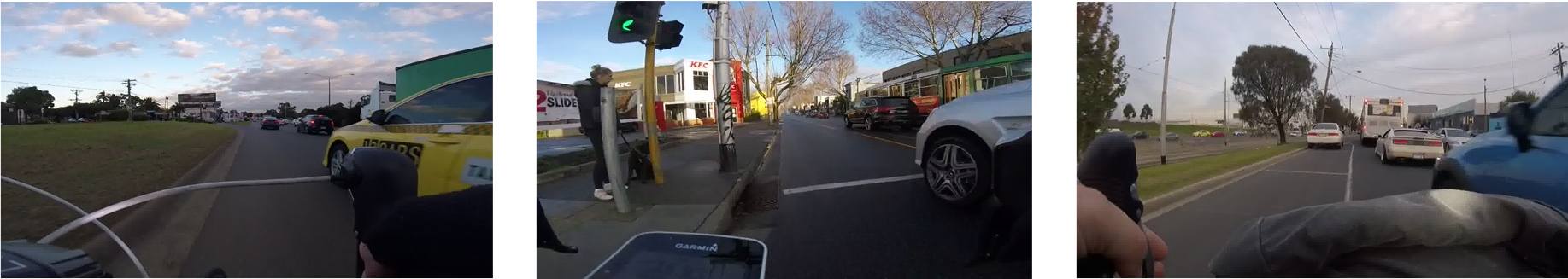}
\caption{Three samples of CP events extracted from the Victorian On-road Cycling Dataset. Video recordings capturing CP events invariably capture additional objects, such as pedestrians, vehicles, and obstacles.}
\label{fig:cpnm_samples}
\end{figure*}

In light of the frequent interactions between cyclists and motor vehicles~\cite{beck2017road}, detection of unsafe interactions has become an important focus for improving cycling safety~\cite{rudolph2022too}. Cycling near miss events can be categorized into distinct types~\cite{ibrahim2021cycling}, of which \emph{close passes}, where a vehicle passes a bike while driving very close to the bike, is the most prevalent~\cite{beck2019much,ibrahim2021cycling,aldred2016cycling}. Quantitative analysis of such events can assist in informing road safety policy, implementing bicycle infrastructure that enhances safety and perceived comfort for cyclists, and assisting cyclists with route planning to avoid potential conflicts and collisions~\cite{ibrahim2021cycling,li2023two,alsaleh2021markov}. In general, close passes can be detected through a variety of bike-mounted sensors, such as video cameras~\cite{ibrahim2021cyclingnet}, distance sensors~\cite{beck2019much}, GPS and mobile motion sensors. In this paper, we focus on close pass (CP) detection from video streams for two reasons: first, video data can be easily and routinely collected by numerous cyclists through the utilization of GoPro or mobile phone cameras. This ease of data collection provides a substantial foundation for empirical investigation; second, the richness of video data offers an enhanced scope for quantitative analysis since attributes of the road environment can be observed from video data.

Recent works~\cite{ibrahim2021cyclingnet,lehtonen2016evaluating,vansteenkiste2016hazard} have treated CP detection as a video classification problem. In early work, researchers invited experienced cyclists to watch videos and identify the CP~\cite{vansteenkiste2016hazard}. Building on the great success of deep learning techniques in computer vision tasks~\cite{ResNet,DenseNet,i3d}, Ibrahim~\textit{et al.}~\cite{ibrahim2021cyclingnet} proposed Cycling-Net to automatically recognize whether the given video clip contains a close pass or not. Despite achieving satisfactory performance, research on CP detection is hindered by two main challenges. Firstly, deep neural networks are usually highly data-hungry, and labeling annotations for CP is a costly and laborious task. For example, in \cite{ibrahim2021cyclingnet}, ground truth CPs are manually labeled by considering the distinct road environments that arise from various traffic policies and infrastructure, which makes it difficult for other researchers to create their own datasets for training models. Secondly, existing CP detection works, such as those in \cite{ibrahim2021cyclingnet,lehtonen2016evaluating}, lack interpretability. Although a video recording detecting CPs tends to capture multiple entities, such as pedestrians, cyclists, and different types of vehicles (refer to Figure~\ref{fig:cpnm_samples}), existing works are trained end-to-end to detect CP without specifying which object led to the final decision. This lack of explainability may reduce the trustworthiness to both cyclists and policy-makers and impede the widespread adoption of CP detection technology.

To overcome these limitations, this paper presents a new benchmark, Cyc-CP, for close pass event detection from video streams. This benchmark is designed to explore two distinct problem formulations for CP detection: scene-level and instance-level detection, respectively. In alignment with preceding studies~\cite{ibrahim2021cyclingnet,lehtonen2016evaluating}, scene-level detection is a classification task, wherein algorithms are tasked with determining the presence of a CP event within a specified video clip. Contrary to scene-level detection, instance-level detection requires the algorithm to pinpoint the object responsible for the CP, thereby enhancing the explainability and utility of the detection process. However, such interpretability hinges on the availability of high-quality labeled data. To collect those training data efficiently, we construct a synthetic dataset in the CARLA~\cite{dosovitskiy2017carla} simulator, wherein a camera is affixed to the front of a bicycle to simulate a diverse array of CP scenarios. This synthetic dataset facilitates the straightforward extraction of critical physical information regarding each passing object (\textit{e.g.}, shape, relative location, and speed), which proves challenging to obtain within real-world settings. 

\begin{figure}[t]
\centering
\includegraphics[width=\textwidth]{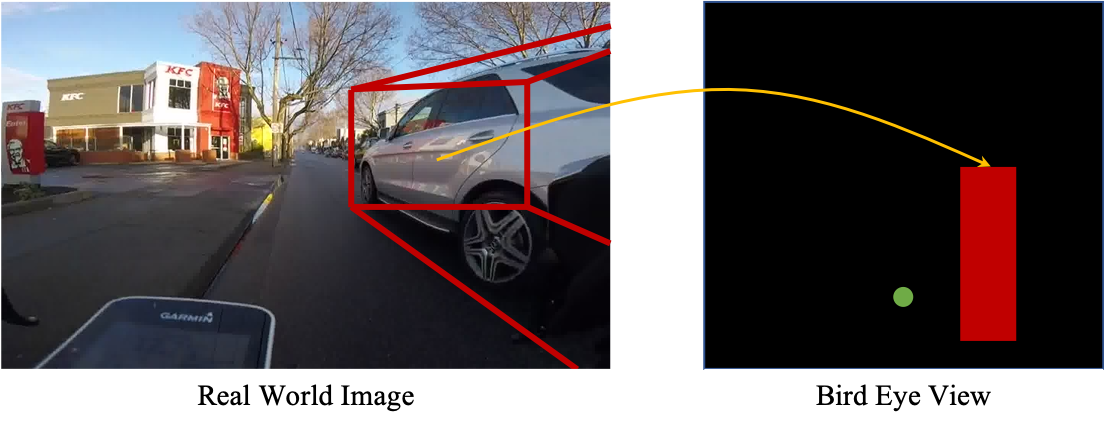}
\caption{Instance-level CP detection involves the detection of objects and subsequent prediction of their attributes such as actual sizes, categories, and 3D locations. CP events are then recognized based on these predicted attributes. The detected objects are displayed as red boxes in a bird's eye view, with the cyclist represented as a green circle.}
\label{fig:motivation}
\end{figure}

In this work, we propose \mj{four} benchmark models for each problem formulation. Specifically, for scene-level detection, we adopt typical video action recognition models, such as I3D~\cite{i3d} and CNN+LSTM~\cite{ResNet,hochreiter1997long}, to predict a Boolean value for each given video clip. Then we employ classification metrics (\textit{i.e.} F1 score, accuracy, and IoU) to measure those models' detection capabilities. \mj{With the recent advancements in large multimodal models (LMMs) demonstrating strong video understanding capabilities, such as GPT-4o~\cite{achiam2023gpt}, we also incorporate a state-of-the-art LMM designed for video understanding, InternVideo 2.5~\cite{wang2025internvideo2}. Using a prompt-based approach, we guide InternVideo 2.5 to detect scene-level Close Pass (CP) events, further exploring the potential of LMMs in this task.} For instance-level detection (see Figure.\ref{fig:motivation}), we adopt a monocular 3D detector to predict the actual size (including width, height, and length) and location for each vehicle in the camera coordinate frame. subsequently, we propose a criterion to detect CPs using those interactions. In this study, we use the VicRoads traffic regulations\footnote{\url{https://www.vicroads.vic.gov.au/safety-and-road-rules/road-rules}} concerning minimum passing distance between cyclists and vehicles as our definition of a CP event. Specifically, an illegal pass is identified when a vehicle fails to maintain a distance of at least 1 meter when passing a cyclist in areas where the speed limit is less than or equal to 60 km/h, or a minimum distance of 1.5 meters in areas where the speed limit is greater than 60 km/h. The contributions of our Cyc-CP include:

\begin{itemize}
    \item To the best of our knowledge, this is the first work towards benchmarking CP detection from video streams. By providing this benchmark, we aim to encourage and inspire researchers to develop effective AI algorithms that can enhance cyclist safety and support policy-makers.
    
    \item We propose two problem formulations for detecting cycling close pass events from video streams: scene-level detection and instance-level detection. 

    \item \mj{We evaluate the capability of LMM in understanding CP events through a prompt-based detection approach.}
    
    \item We conduct extensive experiments on both real-world and synthetic data to evaluate various baseline models and present the corresponding benchmark performances.
\end{itemize}

\section{Related Work}

\subsection{Detection of hazardous events while cycling}\label{sec:cnm}
Detection of potentially hazardous events (also known as \emph{near miss events}) has recently attracted increasing attention~\cite{ibrahim2021cyclingnet,rudolph2022too,lehtonen2016evaluating,vansteenkiste2016hazard}. In early work, researchers~\cite{aldred2015investigating,fuller2013impact,chaurand2013cyclists} utilized surveys to ask participants to describe their behaviors and feelings about the road conditions. Then data was gathered and utilized to analyze the types of risk factors and understand cycling near miss events. Beck \textit{et al.}~\cite{beck2019much} compiled a real-world cycling dataset consisting of videos and vehicle lateral passing distance measurements. They also mounted a button on the bike to let cyclists record moments when passing vehicles made them feel unsafe. The authors conducted a correlation analysis between several factors, including gender, passing distance, speed zone, and the button-press data~\cite{beck2021subjective}. Their findings indicate that passing distance is the most crucial factor contributing to the perceived safety of cyclists on the road. Consistent with these findings, both Ibrahim \textit{et al.}\cite{ibrahim2021cycling} and Aldred \textit{et al.}\cite{aldred2018predictors} have argued that CP is the most commonly occurring type of incident. Therefore, this paper focuses on detecting CP during cycling and explores the ability of a deep neural network to recognize the dynamic interactions between passing vehicles and cyclists.

Building on the success of deep learning systems in many applications~\cite{li2022cross,zhang2020gis,li2022video,li2019knowledge,liu2024context,li2024contrastive,li2023dynamic}, Ibrahim \textit{et al.}~\cite{ibrahim2021cyclingnet} proposed a deep neural network (DNN) for detecting CPs from video data. They collected online videos recorded when people cycle and used crowd-sourced annotation tools to label CP on the videos' timelines. Then, their model is trained on those data for scene-level detection (whether there is a CP event in the given clip or not). In addition to scene-level detection, in this paper, we propose a new problem formulation to detect CP at the instance level, which can improve model interpretability and increase downstream applications. We provide comprehensive evaluations of benchmark algorithms for these two problem formulations.

\subsection{Traffic Conflict Detection}
Traffic conflict is extensively studied~\cite{lehtonen2016evaluating,samerei2021using} using collected videos from CCTV, driving videos~\cite{xia2023automated}, VR-enabled simulation environment~\cite{xu2024analyzing} or surveillance cameras at specific sites such as intersections~\cite{alsaleh2021markov,li2023two}. Those videos usually have fixed views and capture conflicts at specific locations. Lehtonen \textit{et al.}~\cite{lehtonen2016evaluating} invited experienced cyclists to watch video clips and point out near misses manually. In contrast, Samerai \textit{et al.}~\cite{samerei2021using} proposed a binary logistic regression model to cluster different types of conflicts and analyze risk factors automatically. While CP detection from fixed cameras has not attracted a lot of attention in the literature, there is a well-established body of knowledge on the detection of other traffic conflicts from video streams. Researchers employed detection~\cite{pramanik2021real,zhao2021good}, segmentation~\cite{wan2022edge} and tracking~\cite{pramanik2021real,zhang2021v} techniques to explore interactions between two objects and then predict traffic conflicts. However, directly deploying those techniques in our tasks may not achieve promising performance due to two aspects. Firstly, CP detection concerns interactions between cyclists and other passing objects. However, when the camera is mounted on the bicycle, the bicycle is not visible in the collected videos, unlike in overhead fixed-camera data streams. Secondly, our camera is moving along with the bicycle. The depth information of each pixel continuously changes and makes it challenging to estimate the relative locations of detected objects. Therefore, in this paper, we propose a benchmark model based on monocular 3D object detection algorithms for instance-level detection to eliminate the limitations of algorithms that rely on static overhead views.

\subsection{Video Action Recognition}
Detecting close passes of cyclists in real-world settings from video footage is analogous to video action recognition tasks, with the main difference being that while action recognition models identify patterns between objects or for a single visible object, CP tasks require models to identify the interactions between captured objects and an object (with a mounted camera) not shown in the scene. Action recognition from video streams has been extensively studied. The SOTA deep learning-based systems can be categorized by the approach taken to extract features. Some models use spatial features for classification, while others use both temporal and spatial features. For example, Simonyan and Zisserman~\cite{simonyan2014two} introduced a two-stream convolutional structure that exploits both RGB data and optical flow. Wang \textit{et al.}~\cite{wang2015action} developed deep convolutional descriptors based on trajectory pooling. Zheng \textit{et al.}~\cite{zheng2020hybrid} created a hybrid model of the convolutional structure and long-short term memory (LSTM) blocks. Joao and Andrew~\cite{i3d} combined spatio-temporal information through the use of both RGB and optical flow inputs to further improve video action recognition performances. Recently, Arnab \textit{et al.}\cite{arnab2021vivit} presented a pure transformer-based~\cite{vaswani2017attention} model that processes video clips directly, demonstrating the model's effectiveness in capturing both spatial and temporal relationships without convolutional layers. These models have achieved high accuracy on benchmark datasets~\cite{2012UCF101}, which could potentially be used as baseline models to classify and locate close pass captured on video. \mj{With the advancement of LMMs~\cite{achiam2023gpt, liu2024deepseek, wang2025internvideo2}, which exhibit strong video understanding capabilities, these models can further enhance video recognition tasks, leveraging prompt-based approaches to improve CP event detection.} In this paper, we focus solely on the impact of different types of visual features on CP detection, without delving into the feature extraction capabilities of the model.

\subsection{Monocular 3D Object Detection}
For the instance-level formulation, the passing vehicle must be identified. A similar problem in computer vision is monocular 3D object detection~\cite{geiger2013vision, caesar2020nuscenes}. Monocular 3D object detection algorithms estimate an object's 3D position and orientation using a single-view RGB image. These algorithms serve as a fundamental technique in the field of autonomous driving.  Generally, those methods can be categorized into multi-stage and end-to-end approaches. Multi-stage methods \cite{manhardt2019roi,li2019gs3d} typically utilize a 2D object detection model (e.g., Faster-RCNN\cite{ren2015faster} and YOLO~\cite{redmon2016you}) as a backbone to extract 2D object information, which is then projected to a 3D bounding box. Since monocular images lack depth information, Xu \textit{et al.}~\cite{xu2018multi} leveraged the predicted depth information to improve the second stage effectiveness. On the other hand, end-to-end approaches~\cite{wang2021fcos3d,mousavian20173d,peng2019pvnet} aim to directly return the 3D information and pose parameters of the camera~\cite{kim2021survey}, avoiding the nonlinear space for rotation regression and improving efficiency. In this paper, we adopt FCOS3D~\cite{wang2021fcos3d} as the backbone for instance-level CPNM detection.

\section{Cyc-CP Benchmark}
In this section, we first define the scene-level and instance-level CP detection problems and introduce their respective notations. Then, we elucidate details of our proposed benchmark models and other baseline models for both problems.


\begin{figure*}[t]
    \centering
    \includegraphics[width=0.95\textwidth]{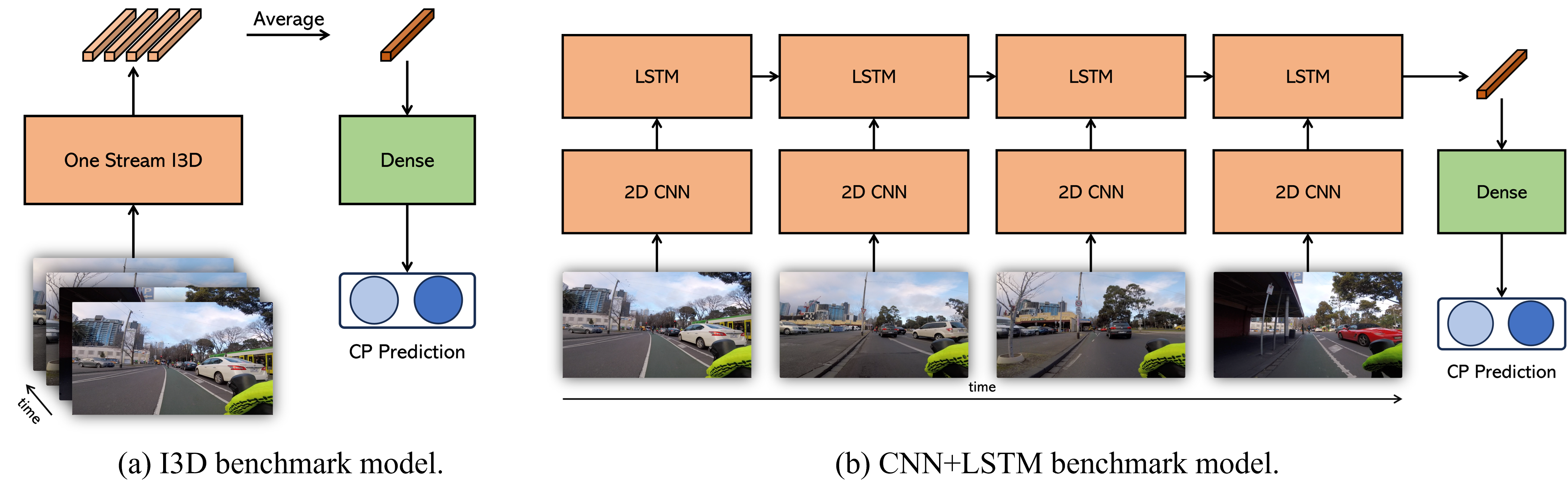}
    \caption{Architectures of the benchmark models for scene-level CP detection.}
    \label{fig:scene level arch}
\end{figure*}


\subsection{Scene-level CP Detection}

\subsubsection{Problem Formulation}

In this paper, we define scene-level CP detection as a classification task, where the objective is to ascertain the presence or absence of a CP event within a provided video clip.  In contrast to previous works~\cite{ibrahim2021cyclingnet,samerei2021using,ibrahim2021cycling} that use subjective observer generated labels, we leverage the local road rules from VicRoads\footnote{\url{https://www.vicroads.vic.gov.au/safety-and-road-rules/road-rules}} (our state-based road rules) to define the CP objectively. A pivotal aspect of this approach is the incorporation of road speed limits into the model's framework, acknowledging the legislative mandate that vehicles must adhere to a minimum passing distance from cyclists—1 meter in areas where the speed limit is 60 km/h or less, and 1.5 meters in zones where the speed limit exceeds 60 km/h. We manually provide the system with speed limit information. This information is captured from videos in real-world dataset and generated for synthetic dataset. For parts where the speed limit sign cannot be read, we set it to above 60 km/h. In this manner, we can automatically label the CP ground truth and generate reproducible experimental results. Given a video clip $X = \{x_1,x_2,\dots,x_l\}$ with RGB channels, a scene-level CP detection algorithm outputs a single CP label $y^*$ for the entire video clip, where $x_i$ refers to the $i$-th frame in a video clip $X$ and $l$ is the length of a video clip. In this paper, we set a uniform length $l=50$ for all video clips.

\subsubsection{Benchmark Model}
The key concept of scene-level CP detection is to recognize risky CP events. Accordingly, temporal information should be extracted for action recognition while spatial features are critical for judging interactions. In this paper, we adopt \mj{three} video action recognition models as our baseline models. 

\noindent\textbf{Inflated 3D ConvNets} (I3D)~\cite{i3d} has been widely used as the backbone network in diverse video analysis tasks due to its spatio-temporal information extraction capabilities. Different from conventional 3D ConvNets, like C3D~\cite{c3d}, which uses 2D ConvNets with additional kernel dimensions to directly create spatio-temporal representations, I3D inflates 2D ConvNets to 3D ConvNets by endowing all the filters and pooling kernels with an additional temporal dimension. This concept significantly increases the depths of 3D ConvNets and initializes 3D ConvNets' parameters with pretrained 2D ConvNets. The basic module in I3D is implemented following Inception-V1~\cite{ioffe2015batch}, in which the spatial and motion features before the last average pooling layer of Inception-V1 are passed through a $3\times3\times3$ 3D convolutional layer with 512 output channels, followed by a $3\times3\times3$ 3D max-pooling layer and through a final fully connected layer. The architecture of our adopted I3D is shown in Figure.\ref{fig:scene level arch} (a). Given a video clip $X$, we get the spatio-temporal representations $f_I(x) \in R^{n*1024}$ from the last Inception module in I3D. After an average pooling layer, we utilize a fully connected layer to project the dense vectors to $y^* \in R^2$ as the CP prediction. Notably, the original I3D is a two-stream network whose inputs are RGB frames and optical flows. In this work, we experiment with three types of input for I3D, namely single-stream feature extraction with only either RGB frames or fused frames~\cite{ibrahim2021cyclingnet}, and two-stream feature extraction with both RGB frames and optical flows. 

\noindent\textbf{CNN+LSTM} consists of a 2D ConvNets, such as ResNet~\cite{ResNet} and DenseNet~\cite{DenseNet}, and a recurrent layer on top, such as LSTM~\cite{hochreiter1997long}.  The architecture of our adopted CNN+LSTM model is shown in Figure.\ref{fig:scene level arch} (b). We reuse a pre-trained 2D ConvNets, ResNet101~\cite{ResNet}, to extract features independently from each frame. Then we position an LSTM layer to encode state and capture temporal ordering and long-range dependencies after the last average pooling layer of ResNet. In the end, a fully connected layer is added on top of the outputs of the last LSTM layer for the CPNM prediction $y^*$. Existing work CyclingNet~\cite{ibrahim2021cyclingnet} follows this architecture but its inputs are the sum of RGB representations and optical flows. Similar to that for I3D, in this work, we experiment with three types of input for CNN+LSTM as well, namely single-stream feature extraction with only either RGB frames or fused frames~\cite{ibrahim2021cyclingnet}, and two-stream feature extraction with both RGB frames and optical flows. 

\noindent\textbf{Large Multi-modal Models}~\cite{wang2025internvideo2} \mj{have demonstrated strong video understanding capabilities by effectively integrating spatial and temporal reasoning. Unlike traditional video recognition models that are trained on specific tasks, these models leverage multi-task multi-modal learning to process both visual and textual inputs, enabling more context-aware interpretations of video content. However, to achieve accurate and reliable detection, LMMs require prompt-based guidance to focus on specific tasks.}

\begin{figure*}[th!]
\centering
\includegraphics[width=\textwidth]{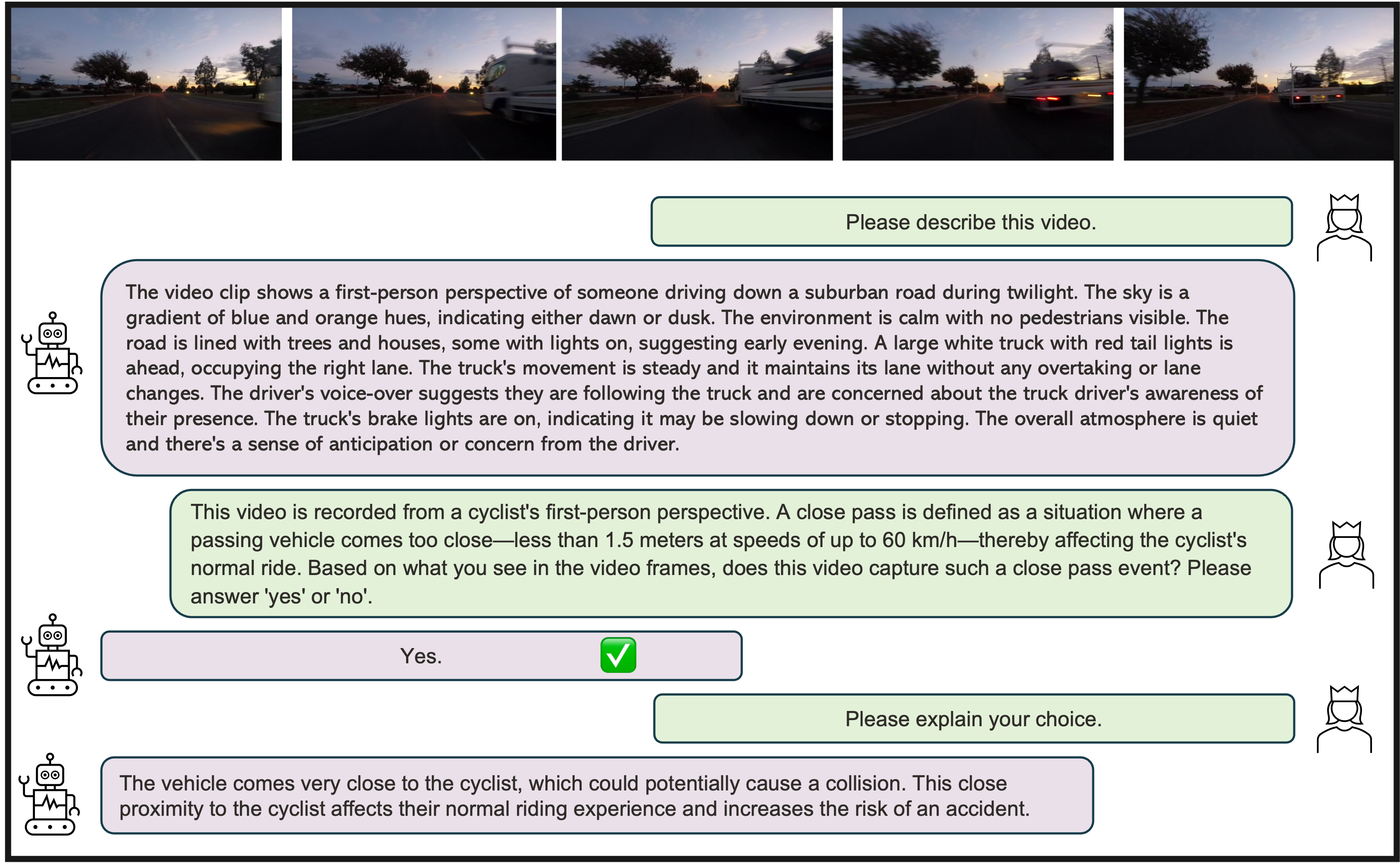}
\caption{\mj{Illustration of adapting a large multi-modal model for scene-level CP detection. We asked the model to identify a CP event and explain its choice step by step.}}
\label{fig:llm}
\end{figure*}

\mj{To leverage VLMs for CP detection, we first prompt the model to generate a natural language description of the given video clip. This step enhances its contextual understanding by encouraging the model to capture relevant details about the scene, such as road conditions, surrounding vehicles, and cyclist interactions. Next, we provide the model with a clear definition of a CP event and explicitly ask whether the given clip contains a CP instance. Finally, we instruct the model to provide a justification for its decision, explaining the reasoning behind its classification. We present this process including all the detailed prompts in Figure~\ref{fig:llm}. This multi-step prompt-based approach not only improves detection accuracy but also enhances interpretability, allowing us to understand how the model arrives at its conclusions. By combining scene description, definition-based classification, and justification, our method enables a more transparent and explainable CP detection process, making it more suitable for real-world applications in cycling safety.}

\subsubsection{Objectives}

For both I3D and CNN+LSTM networks, we use the commonly used cross-entropy loss for CP prediction:

\begin{align}
    L = -\frac{1}{N}\sum_{i}(y_i*log(y^*_i)+(1-y_i)*log(1-y^*_i))
\end{align}
where $y_i$ represents the ground truth of a given video clip, $y_i=1$ means there is the referring video clip contains a CP event.

\begin{figure}[t!]
\centering
\includegraphics[width=\textwidth]{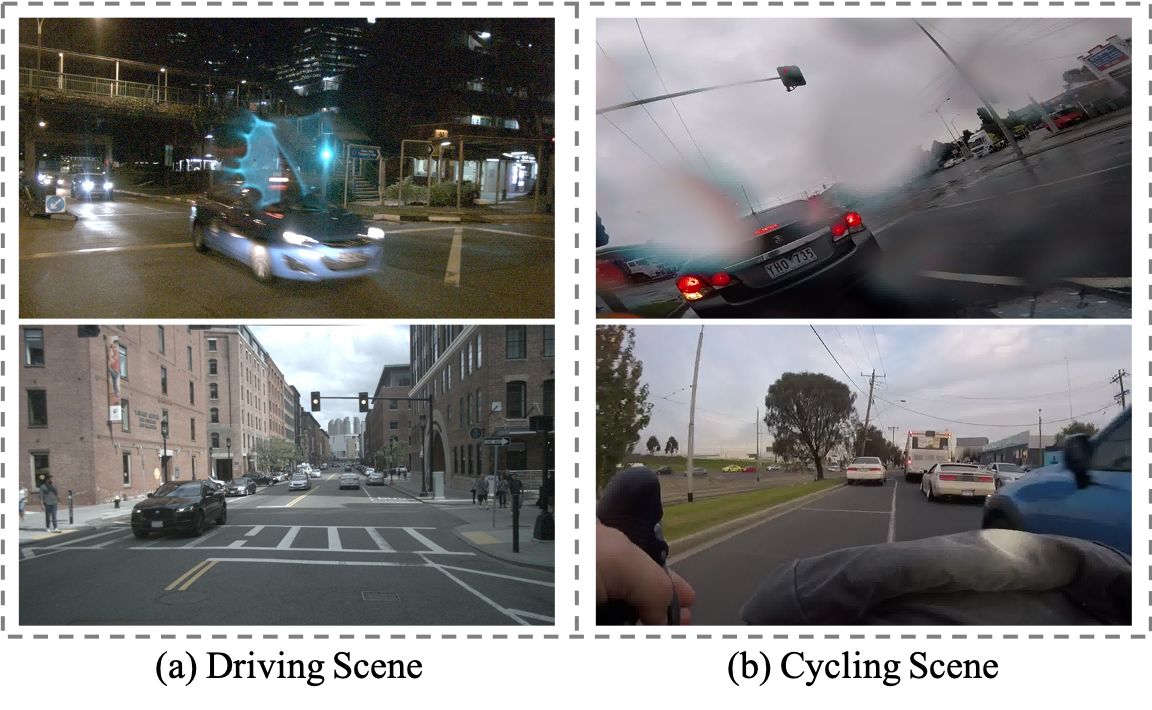}
\caption{Comparison of videos from NuScenes driving~\cite{nuScenes} and our Victorian On-road Cycling scenes. Videos recorded by cameras that are mounted on bicycles facing forward are usually wobbly and truncated while driving scenes are more stable.}
\label{fig:challenges}
\end{figure}
\begin{figure*}[t!]
\centering
\includegraphics[width=0.93\textwidth]{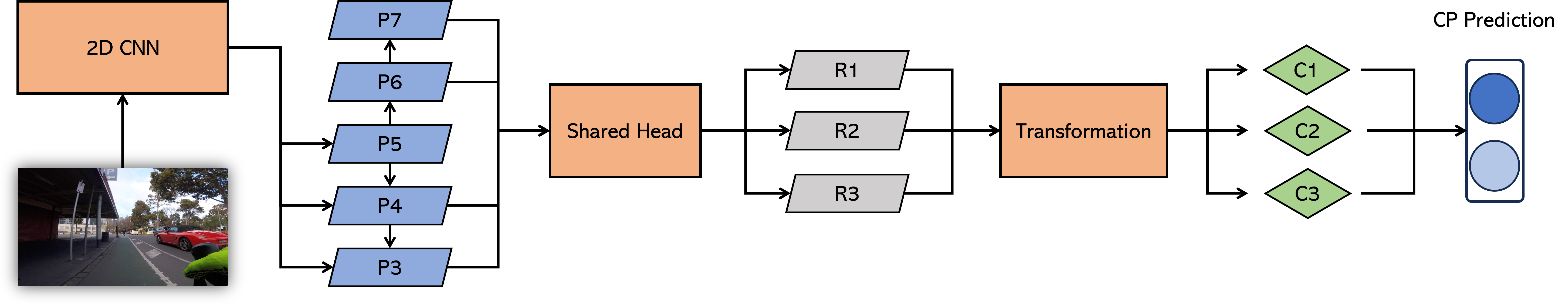}
\caption{The framework of our proposed ICD for instance-level cycling near miss detection. P3 to P7, R1 to R3, and C1 to C3 represent feature maps from levels 3 to 7, three kinds of regression targets, and our proposed three criteria, respectively.}
\label{fig:icd}
\end{figure*}

\subsection{Instance-level CP Detection}

\subsubsection{Problem Formulation}
When a video clip includes multiple vehicles, scene-level detection algorithms do not disambiguate which vehicle was involved in the CP event. To provide the ability for more fine-grained analysis, we propose a new CP detection problem formulation, instance-level detection. For instance-level detection, the algorithm must identify which vehicle causes the CP event. To achieve this goal, a model must detect all passing vehicles and then characterize their interactions with the cyclist. As before, we define a CP by referring to our local road rules. The instance-level CP detection algorithm, when applied to a video clip $X$, has the capability to identify and categorize various objects within the scene. For each detected object, the algorithm is designed to predict specific information about that object, which includes the object's category $c_i$, 3D size (width $w_i$, height $h_i$ and length $l_i$), and 3D center of their bounding boxes (relative location to the camera, consisting of forward-distance $d_f$, right-distance $d_r$, vertical distance $d_v$ and allocentric orientation angle $\theta$). After detecting objects and extracting the above information, the algorithm employs a set of criteria, denoted by $f_c(\cdot)$, to predict whether a CP event is likely to occur. This process involves analyzing the collected data for each object to determine its role in a CP. The whole process can be formulated as:

\begin{align}
f_c(c_i, w_i, h_i, l_i, d_f, d_r, d_v, \theta)  = 
\begin{cases} 
1 & \text{if passes the criteria} \\
0 & \text{otherwise}
\end{cases}
\label{eq:ICD_prediction}
\end{align}

The task of estimating objects' information is similar to the monocular 3D object detection (M3OD) task~\cite{geiger2013vision}. However, instance-level CP detection tasks present unique challenges. Firstly, the use of cameras that are mounted on bicycles facing forward leads to a dynamic camera coordinate system and produces oblique or wobbly videos (see Figure.~\ref{fig:challenges}). Although ConvNets can still identify distorted objects due to translation invariance, their ability to accurately regress the actual size and location of objects may be degraded compared to pre-trained models on commonly used M3OD benchmarks. Secondly, cameras mounted on bicycles may capture truncated vehicles when nearby, which is critical to detecting close by vehicles. However, such samples constitute only a small portion of publicly available M3OD datasets. In contrast, cameras mounted on cars often capture stable videos and observe vehicles that are further away, thus providing a more favorable environment for object detection.

\subsubsection{Benchmark Model}

\begin{figure*}[t!]
    \centering
    \includegraphics[width=0.9\linewidth]{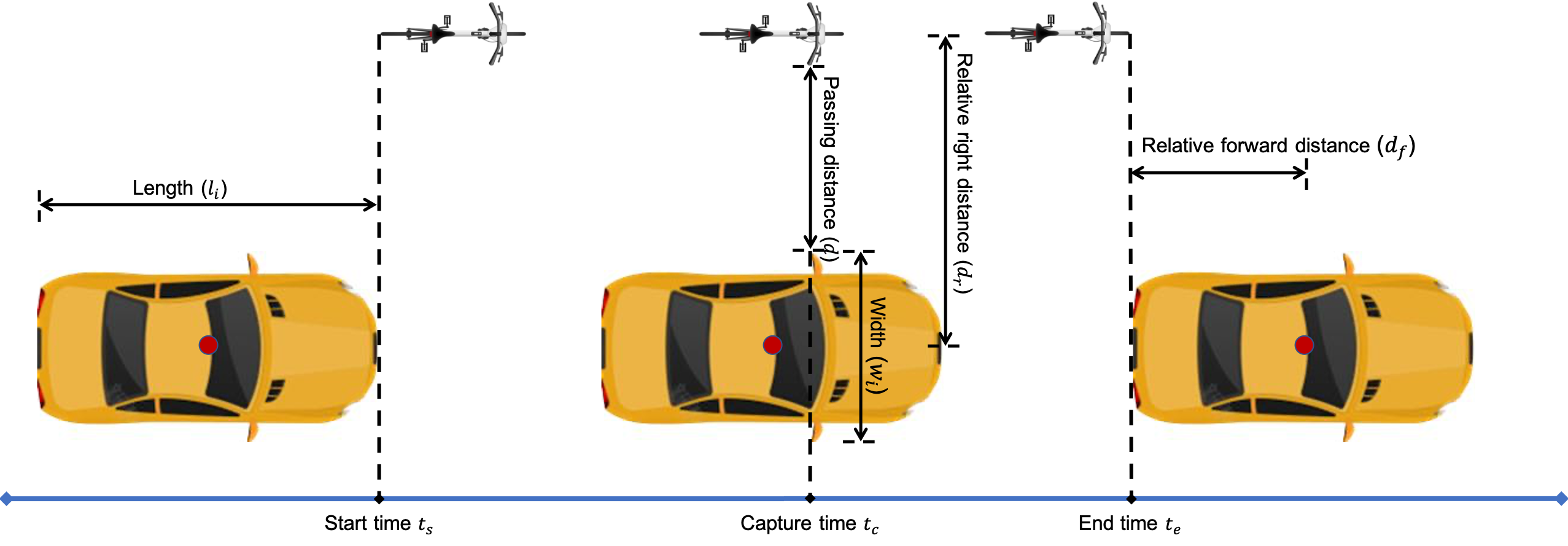}
    \caption{Illustrations of a passing event. For each case, we detect each object and predict its 3D center and 3D size in camera coordinates. The red circle represents the vehicle's center.}
    \label{fig:criteria1}
\end{figure*}

To address those challenges, we propose a two-stage deep neural network, named ICD, for \textbf{I}nstance-level \textbf{C}ycling CP \textbf{D}etection. Our ICD consists of a monocular 3D object detector, FCOS3D~\cite{wang2021fcos3d}, a simple but powerful M3OD network for detecting vehicles, and a post-procedure to recognize CPs. The overview architecture of ICD is shown in Figure.\ref{fig:icd}.

In this paper, we will introduce FCOS3D briefly, more details can be found in the original paper \cite{wang2021fcos3d}. The FCOS3D is a fully convolutional one-stage detector and consists of three components. Firstly, it uses a pre-trained ResNet101~\cite{ResNet,ImageNet} with deformable convolutions~\cite{DeformConv} for feature extraction. Then a Feature Pyramid Network~\cite{FPN} is proposed for detecting objects at different scales. Lastly, different scale features are fed into a shared head which consists of 4 shared convolutional blocks and small heads for 7-DoF regression targets to the 2.5D center and 3D size. Then the 2.5D center can be easily transformed back to 3D space with a camera intrinsic matrix. The 3D center and size are essential ingredients for our proposed ICD to recognize a CPNM via a post-procedure. 

\label{subsubsec:postprocedure_ICD}
Within the \textbf{\textit{post-procedure}}, three criteria are designed according to our local road rules for assessing detected vehicles. Only a vehicle that fulfills all the criteria can be recognized as a CP. Future works can easily adjust criteria to fit local traffic policies by choosing new hyper-parameters. As illustrated in Figure.\ref{fig:criteria1}, for each detected vehicle, we can estimate the passing distance as follows:
\begin{align}
    d = d_r - w_i/2 - d_{ch}
    \label{eq:estimate_passing_distance}
\end{align}
where $d_r$ is the relative right distance between the center of the bike and the car, $w_i$ is the width of the car, and $d_{ch} = 0.5$ is the distance in meters between our mounted camera and the outer-right-most point of the handlebar. According to our local road rules, our \textbf{first criterion} is that $d \geq 1$ when the speed limit is less than 60km/h, or $d \geq 1.5$ otherwise. Then the \textbf{second criterion} makes sure the detected vehicle is passing our cyclist, which can be formulated as:
\begin{align}
    -l_i/2 - l_b \leq d_f \leq l_i/2
    \label{eq:determine_passing}
\end{align}
where $l_b = 1.8$ refers to the length of a bicycle and $l_i$ and $d_f$ are the detected vehicle length and the predicted forward distance respectively in Eq. \ref{eq:ICD_prediction}. Because we are utilizing forward-facing video data for events that occur when the vehicle is not in the camera's field of view, Eq. \ref{eq:determine_passing} not only considers the cases where a vehicle is approaching from behind the bike, i.e., $-l_i/2 - l_b \leq d_f$, but also takes into account the cases where a vehicle is passing or has just passed the bike, i.e., $d_f \leq l_i/2$. In Victoria, motor vehicles drive on the left side of the road, and so overtaking occurs to the right. Therefore, when a cyclist is overtaken by a car (such as when a cyclist is in a painted bikelane), the motor vehicle passes on the right of the cyclist. Therefore, our \textbf{third and last criterion} is $d_r \geq 0$ where $d_r$ is the predicted right distance in Eq. \ref{eq:ICD_prediction}.

\subsubsection{Objectives}

We followed the objectives proposed by \cite{wang2021fcos3d} to train our ICD model. Firstly, to predict the location and orientation of objects, we incorporate a soft binary classifier, center-ness $c$. By doing so, we can identify which points are closer to the center of an object, and suppress low-quality predictions that are far from the center. Thus, in the regression branch, we aim to predict the offsets $\Delta x$, $\Delta y$ from the center to a specific foreground point $x$, $y$, its corresponding depth $d$, the width $w$, length $l$, height $h$, the angle $\theta$, and the velocities $v_x$, $v_y$ of an object, direction class $C_\theta$, and center-ness $c$, while in the classification branch, we output the object category. We define the loss functions for classification and regression targets separately and compute their weighted sum to obtain the total loss. Specifically, the commonly used focal loss~\cite{RetinaNet} is employed for object classification loss. 
\begin{equation}
    \centering
    L_{cls} = -\alpha(1-p)^\gamma log(p)
\end{equation}
where $p$ represents the class probability of a predicted box, we adopt the same configuration as the original paper by setting $\alpha = 0.25$ and $\gamma = 2$ for this task. Regarding attribute classification, we employ a basic cross-entropy loss, denoted as $L_{attr}$. 

For the regression branch, we utilize a smooth L1 loss for every regression target, except center-ness, accompanied by corresponding weights that take into account their respective scales:
\begin{equation}
    \label{eqn: loc_loss}
    \centering
    L_{loc} = \sum_{b\in (\Delta x, \Delta y, d, w, l, h, \theta, v_x, v_y)} SmoothL1(\Delta b)
\end{equation}
where the weight of $\Delta x, \Delta y, w, l, h,$ and $\theta$ error is 1, a weight of 0.2 to $d$, and a weight of 0.05 to $v_x$ and $v_y$, we note that despite employing $exp(x)$ for depth prediction, we compute the loss in the original depth space instead of the log space, as it results in more precise depth estimation, as demonstrated empirically. We employ the softmax classification loss and binary cross-entropy (BCE) loss for direction classification and center-ness regression, respectively, which are denoted as $L_{dir}$ and $L_{ct}$. Lastly, the total loss is:
\begin{equation}
    \centering\footnotesize
    L = \frac{1}{N_{pos}}(\beta_{cls}L_{cls}+\beta_{attr}L_{attr}+\beta_{loc}L_{loc}+\beta_{dir}L_{dir}+\beta_{ct}L_{ct})
\end{equation}
where $N_{pos}$ is the number of positive predictions and $\beta_{cls} = \beta_{attr} = \beta_{loc} = \beta_{dir} = \beta_{ct} = 1$.

\noindent\textbf{Inference} During inference, given an input image, we forward it through the network and obtain bounding boxes with their class scores, attribute scores, and center-ness predictions. We multiply the class score and center-ness as the confidence for each prediction, conduct rotated Non-Maximum Suppression (NMS) into the bird-eye view, and apply the post-procedure introduced in \S~\ref{subsubsec:postprocedure_ICD} to get the final results.

\subsection{Benchmark Datasets}

\subsubsection{NuScenes}

\begin{figure}[h]
\begin{minipage}{0.99\textwidth}
\centerline{\includegraphics[width=\textwidth]{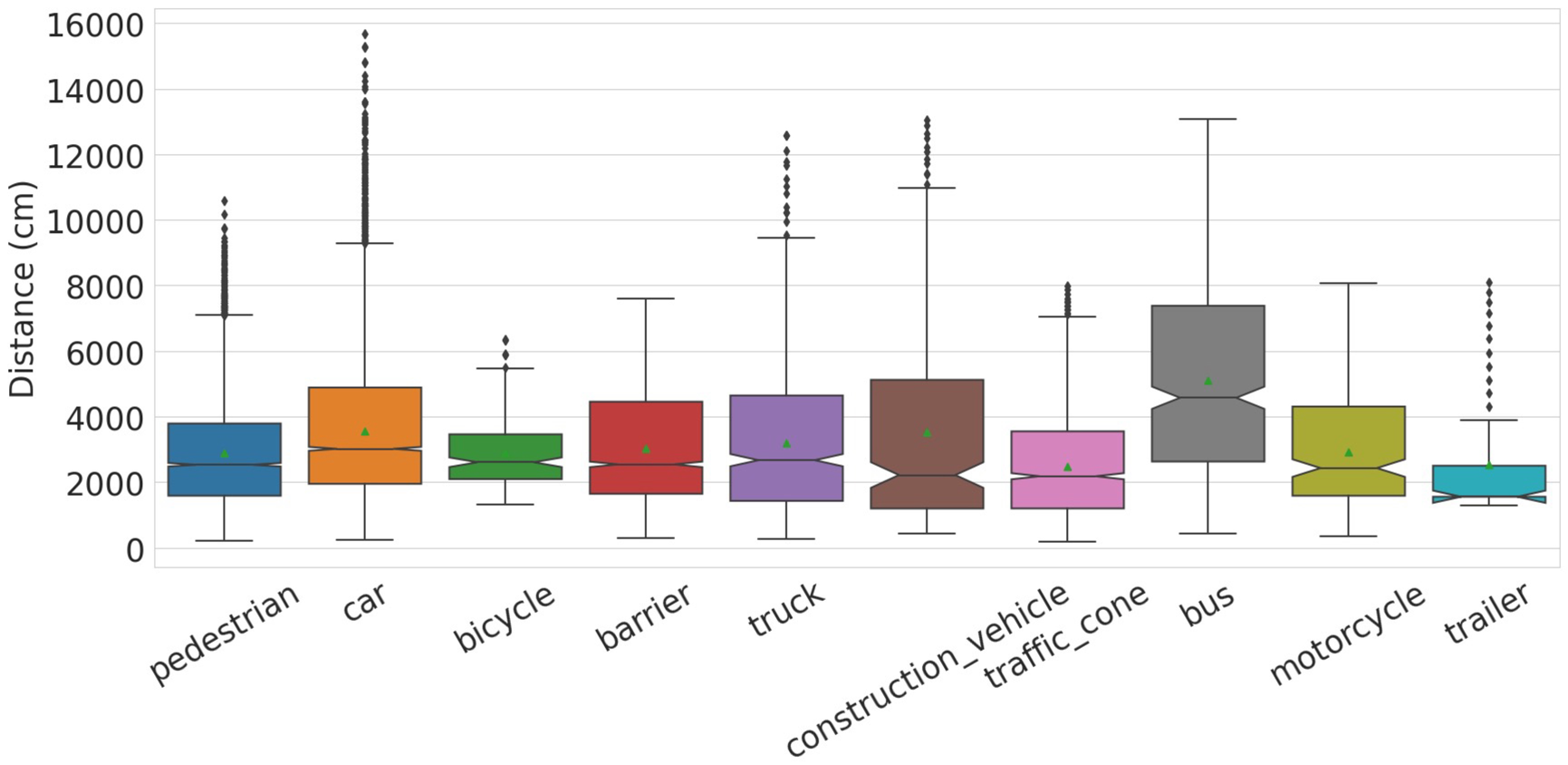}}
\centerline{(a) Euclidean distances from the camera.}
\end{minipage}
\begin{minipage}{0.99\textwidth}
\centerline{\includegraphics[width=\textwidth]{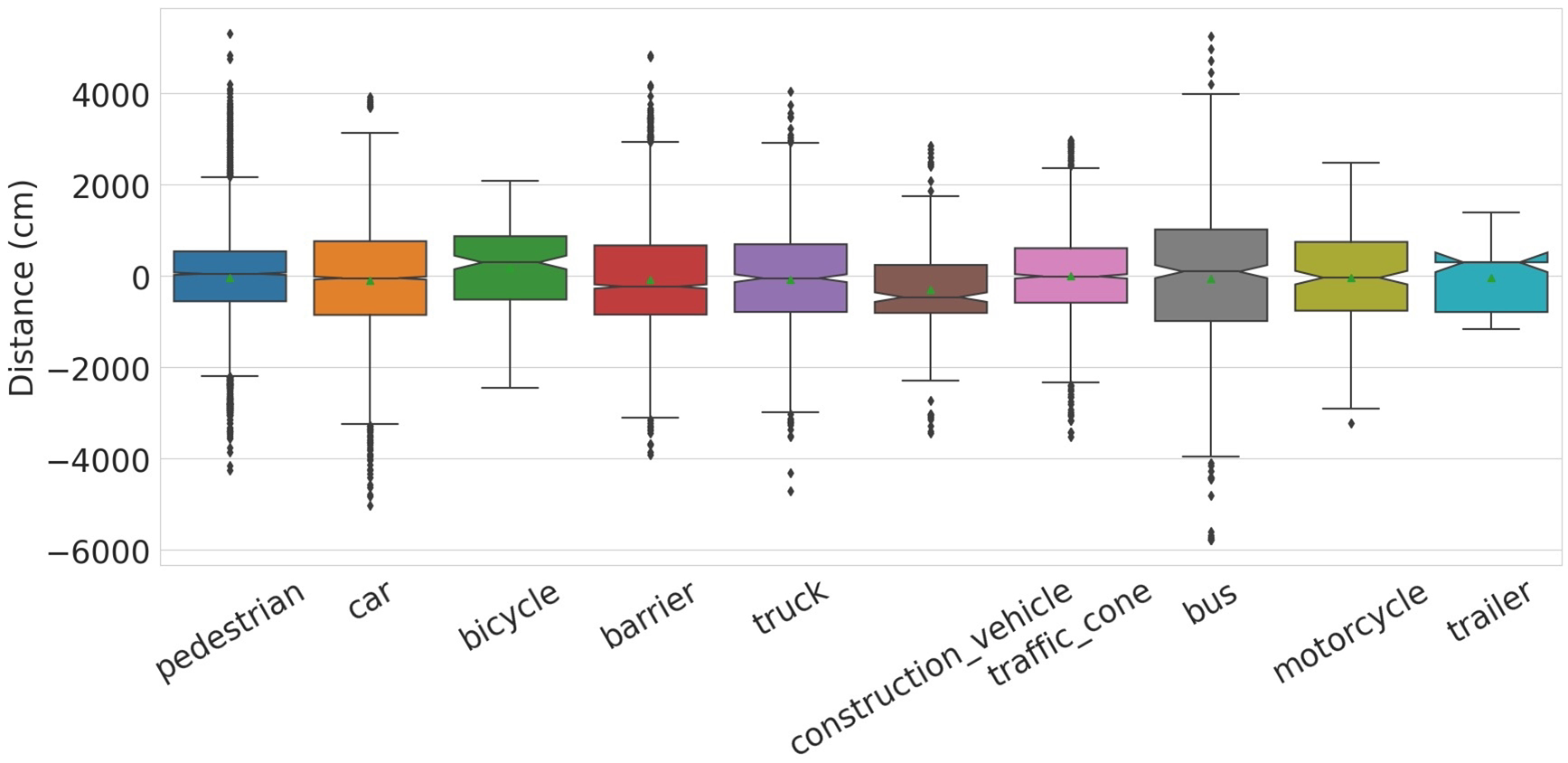}}
\centerline{(b) Right-forward axis distances.}
\end{minipage}
\caption{Statistics of distances between detected object centrenesses and the camera in the NuScenes~\cite{nuScenes} dataset, where green triangles represent mean values.}
\label{fig:nuscenes}
\end{figure}

NuScenes~\cite{nuScenes} is a publicly available multi-modal dataset for autonomous driving. The full dataset provides 1,000 scenes  from North American and Asian locations, resulting in a total of 1,200,000 camera images and more than 140 million objects that are fully annotated with 3D bounding boxes from 10 categories, from which vehicles and bicycles are most related to our task. More details about the NuScenes dataset can be found in ~\cite{nuScenes}, here we visualize the mean distances from the car-mounted camera of 10 classes of objects in Figure.\ref{fig:nuscenes}. Most annotated vehicles are far from the camera in NuScenes (more than 30 meters). Our task focus is on riding scenes instead of driving situations, where close object interactions are of importance. Therefore, we utilize the NuScenes dataset for pre-training, to endow baseline models with capabilities to recognize objects commonly seen in a variety of traffic conditions.

\subsubsection{CARLA Synthetic Dataset}

\begin{figure}[h]
\begin{minipage}{0.49\textwidth}
\centerline{\includegraphics[width=\textwidth]{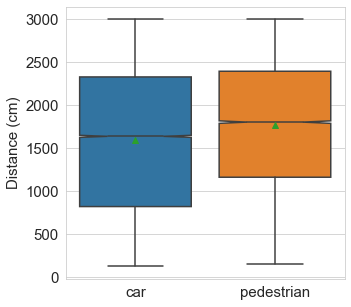}}
\centerline{(a) Actual distances.}
\end{minipage}
\begin{minipage}{0.49\textwidth}
\centerline{\includegraphics[width=\textwidth]{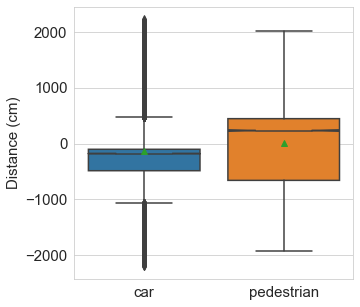}}
\centerline{(b) Right-forward distances.}
\end{minipage}
\caption{Statistic of different kinds of distance between detected objects and camera in our CARLA Synthetic dataset, where green triangles represent mean values.}
\label{fig:carla_statis}
\end{figure}

\begin{figure*}[h]
\centering
\includegraphics[width=0.3\textwidth]{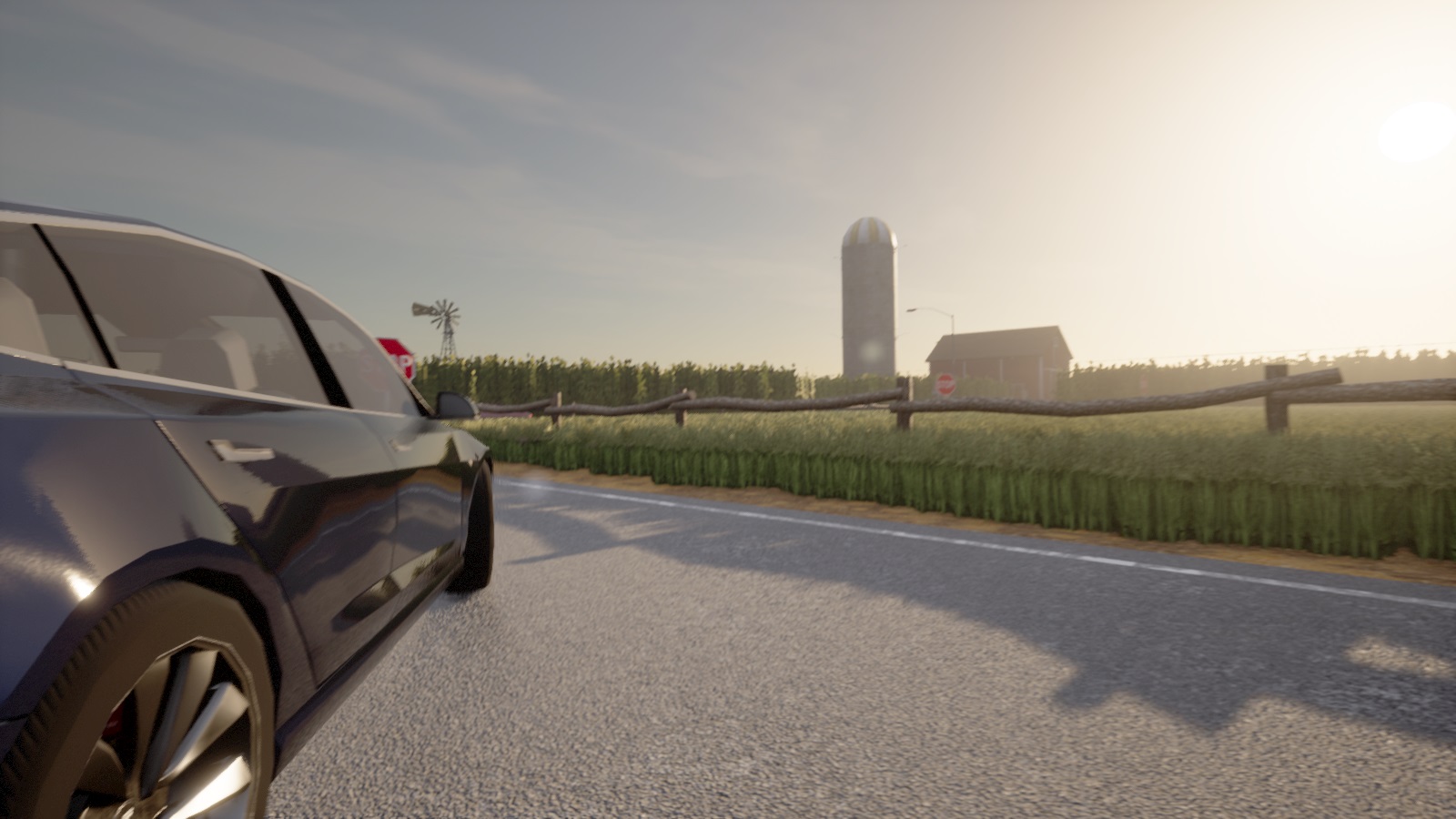}
\includegraphics[width=0.3\textwidth]{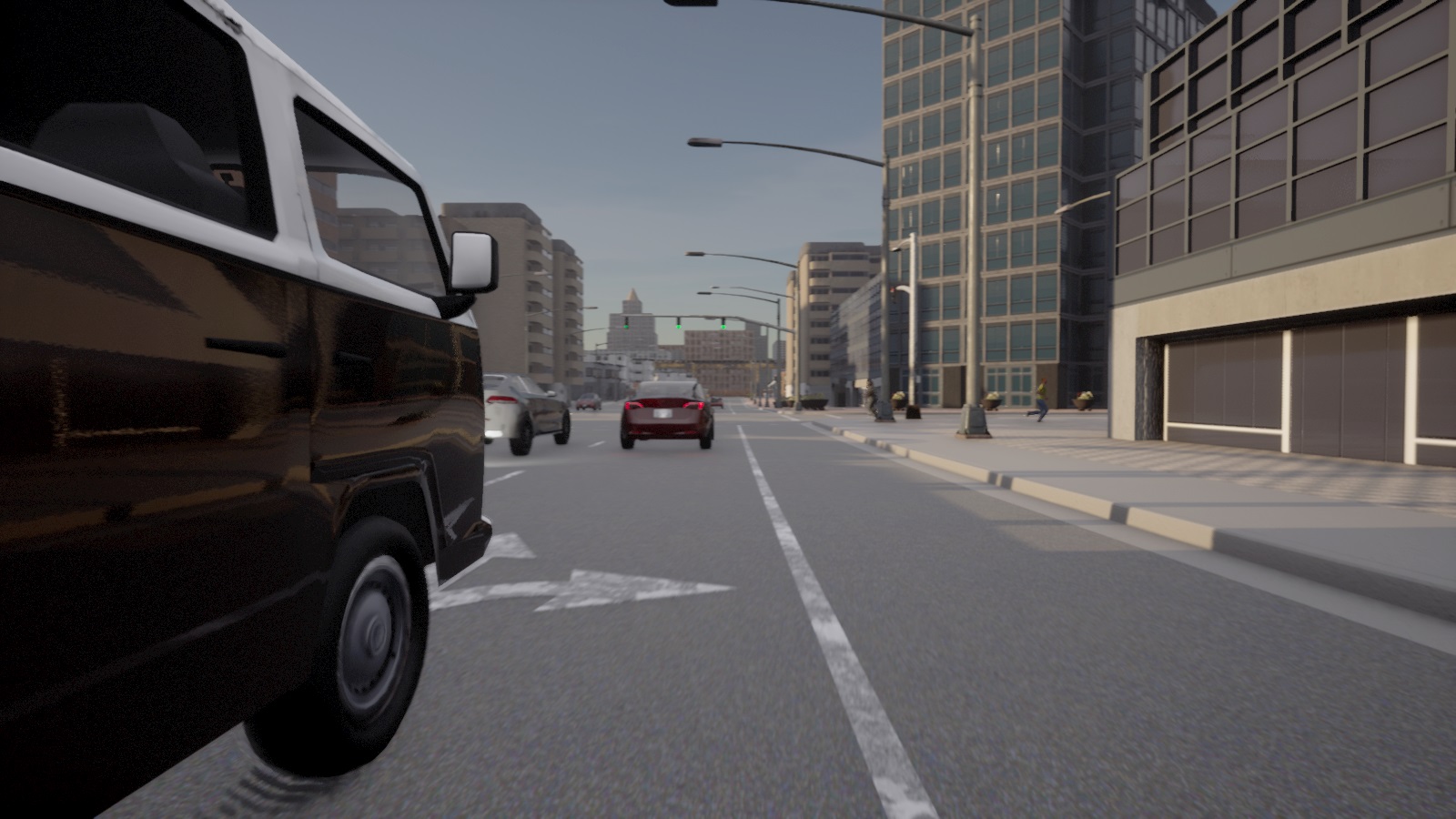}
\includegraphics[width=0.3\textwidth]{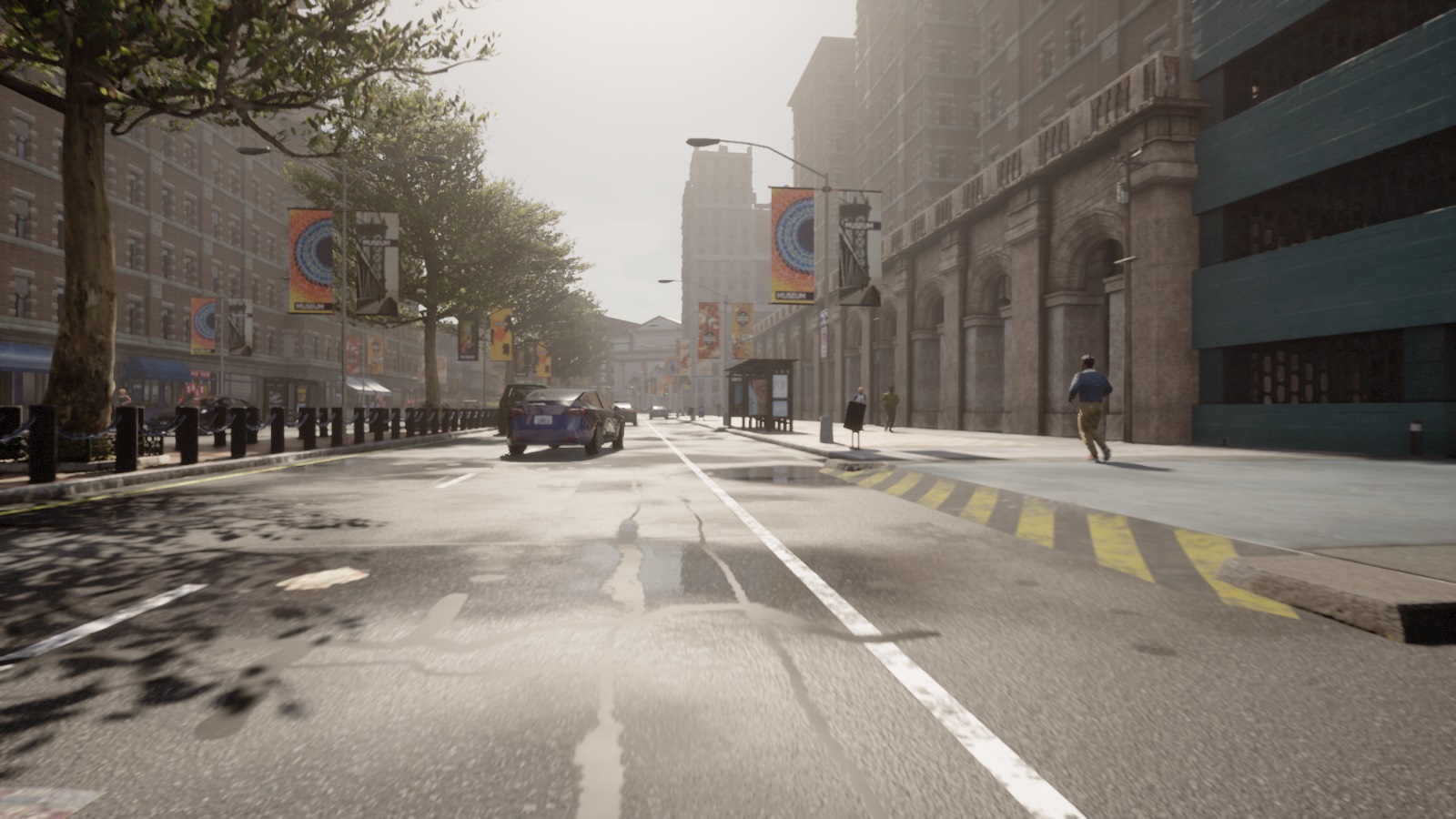}
\caption{Examples of images generated by the CARLA simulator}
\label{fig:carla_images}
\end{figure*}

To simulate riding situations and provide more close interactions, we generate a synthetic CP dataset using the CARLA~\cite{dosovitskiy2017carla} simulator. This simulator provides eight high-fidelity built-in maps set in urban and semi-urban environments. The vehicles, including bicycles, can be generated and controlled using the provided Python Application Programming Interface (API). Using this functionality, we first simulated a series of close-pass scenarios. Consistent with the real-world dataset (described below) and relevant road legislation, we focused on events where a motor vehicle passed the cyclist with a lateral distance of less than 1.5 meters. We generated a bicycle and another randomly chosen vehicle from the supported pool of vehicles available in CARLA in random spawn locations of a given CARLA map. We randomized the parameters such as the distance between the bicycle and the other vehicle, the vehicle speeds, the lateral passing distance, the lateral location of the bicycle and the vehicle in the lane, the time of the day, and the weather. We also generated other vehicles and pedestrians around the cyclist and the interacting vehicle. Examples of generated images are shown in Figure.\ref{fig:carla_images}. Comparing Figure.\ref{fig:carla_statis} and Figure.\ref{fig:nuscenes}, passing vehicles in the simulated dataset are closer to the camera. Fine-tuning models on such a synthetic dataset aims to optimize regression branches.

\subsubsection{Victorian On-road Cycling (VOC) Dataset}

\begin{figure*}[h]
\begin{minipage}{0.45\textwidth}
\centerline{\includegraphics[width=\textwidth]{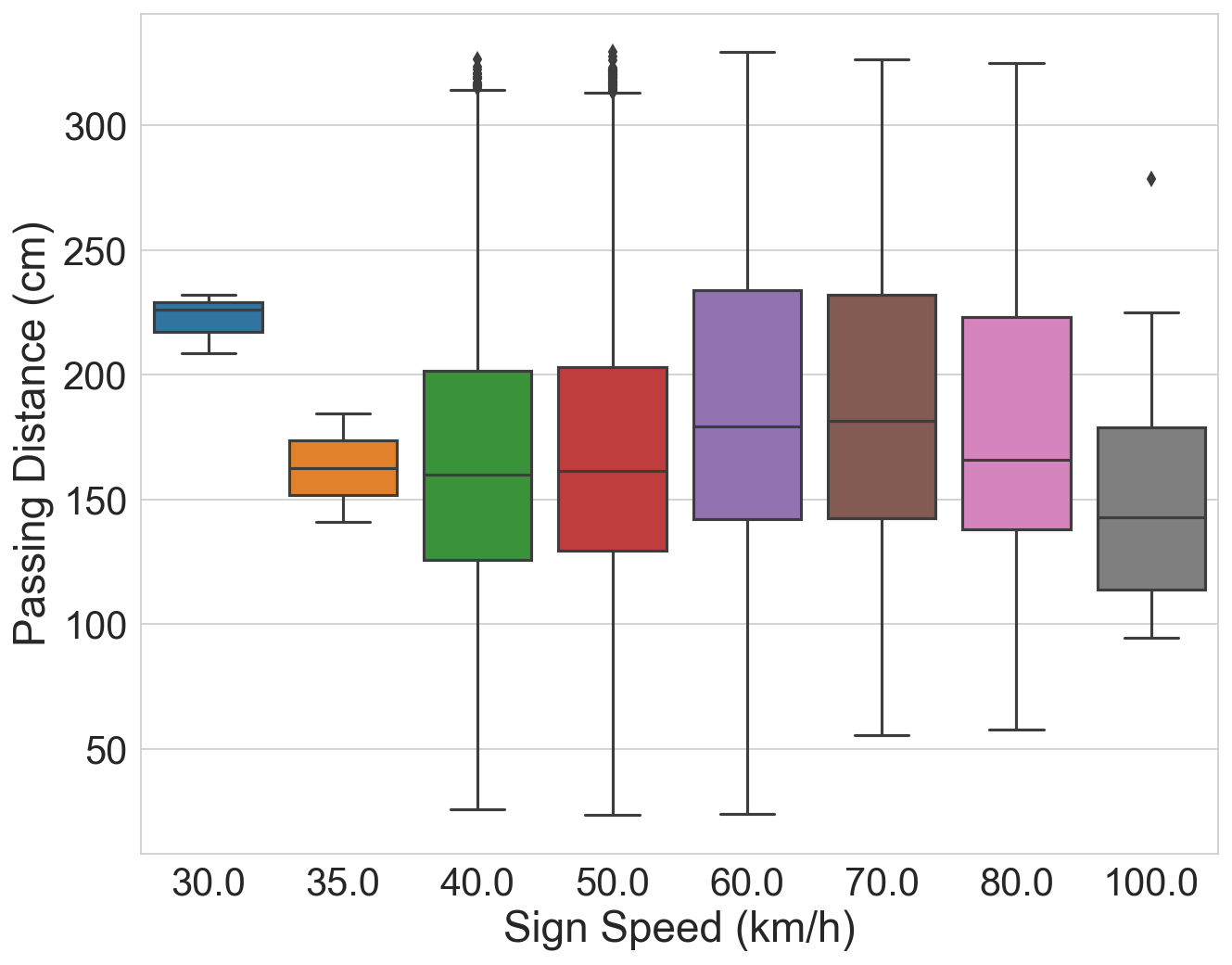}}
\centerline{(a) All passing events.}
\end{minipage}\hfill
\begin{minipage}{0.45\textwidth}
\centerline{\includegraphics[width=\textwidth]{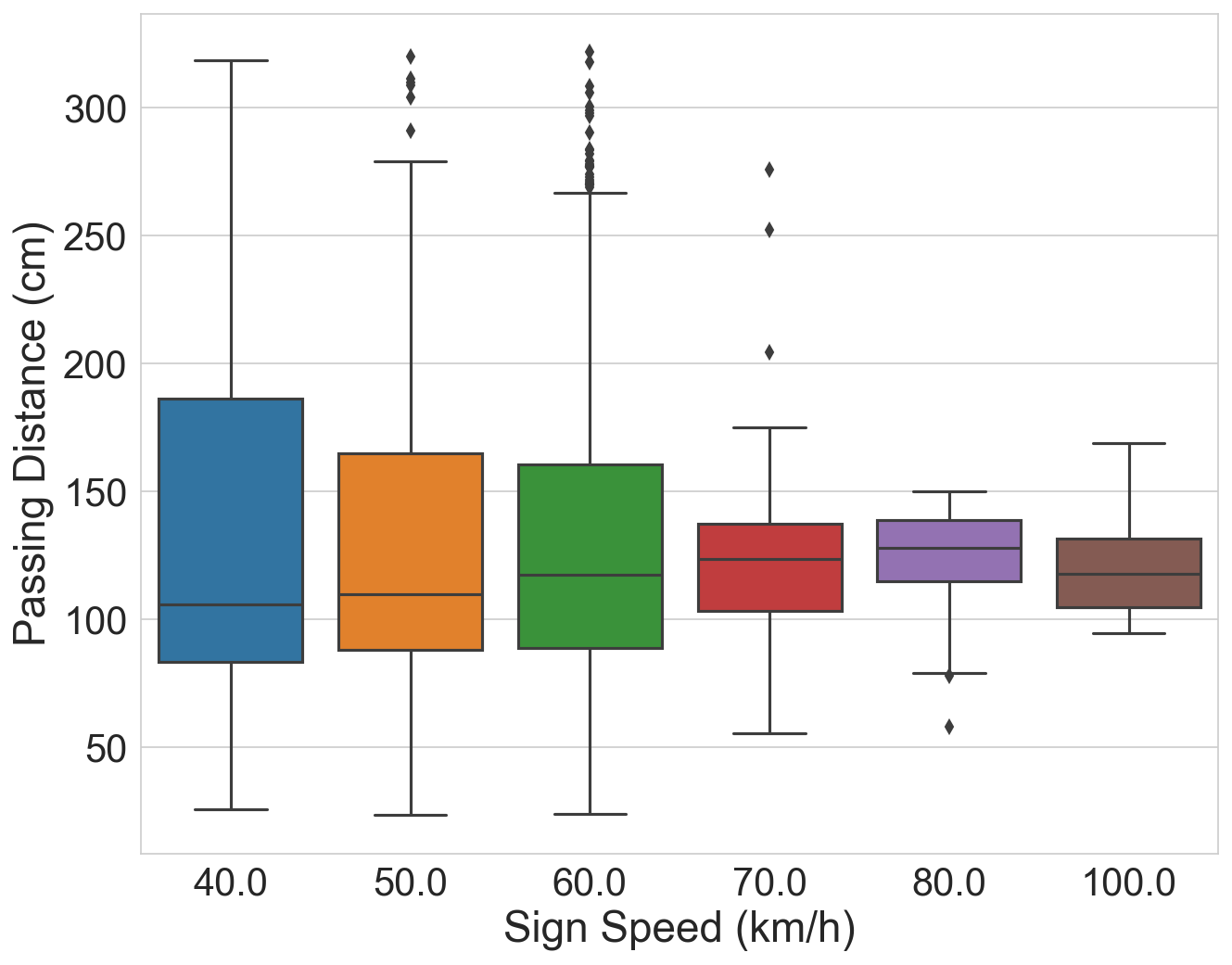}}
\centerline{(b) After negative and positive samples balanced.}
\end{minipage}
\caption{Mean passing distances by limit speeds in our Victorian On-road Cycling dataset.}
\label{fig:VOC_boxplot}
\end{figure*}

\begin{figure}[h]
\includegraphics[width=0.9\textwidth]{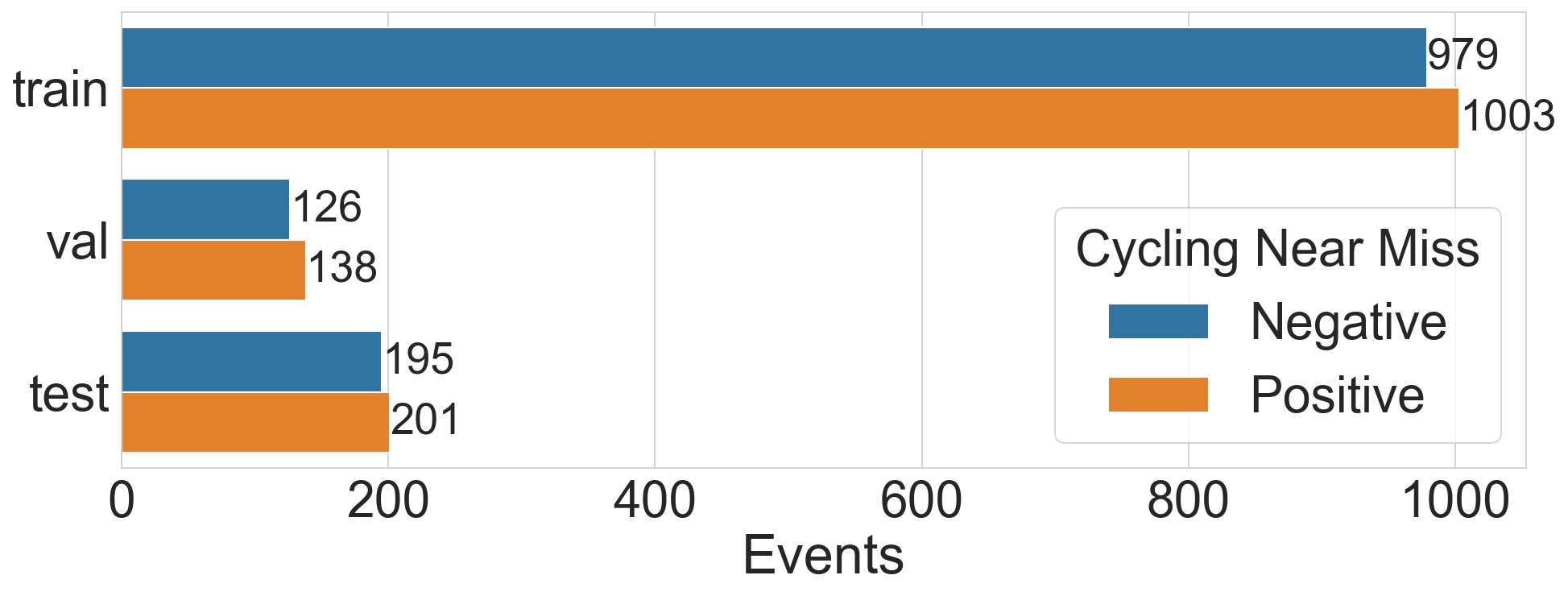}
\caption{Negative and positive cycling near misses in our VOC training, validation, and testing subsets.}
\label{fig:voc_countnm}
\end{figure}

\begin{figure}[h]
\includegraphics[width=0.98\textwidth]{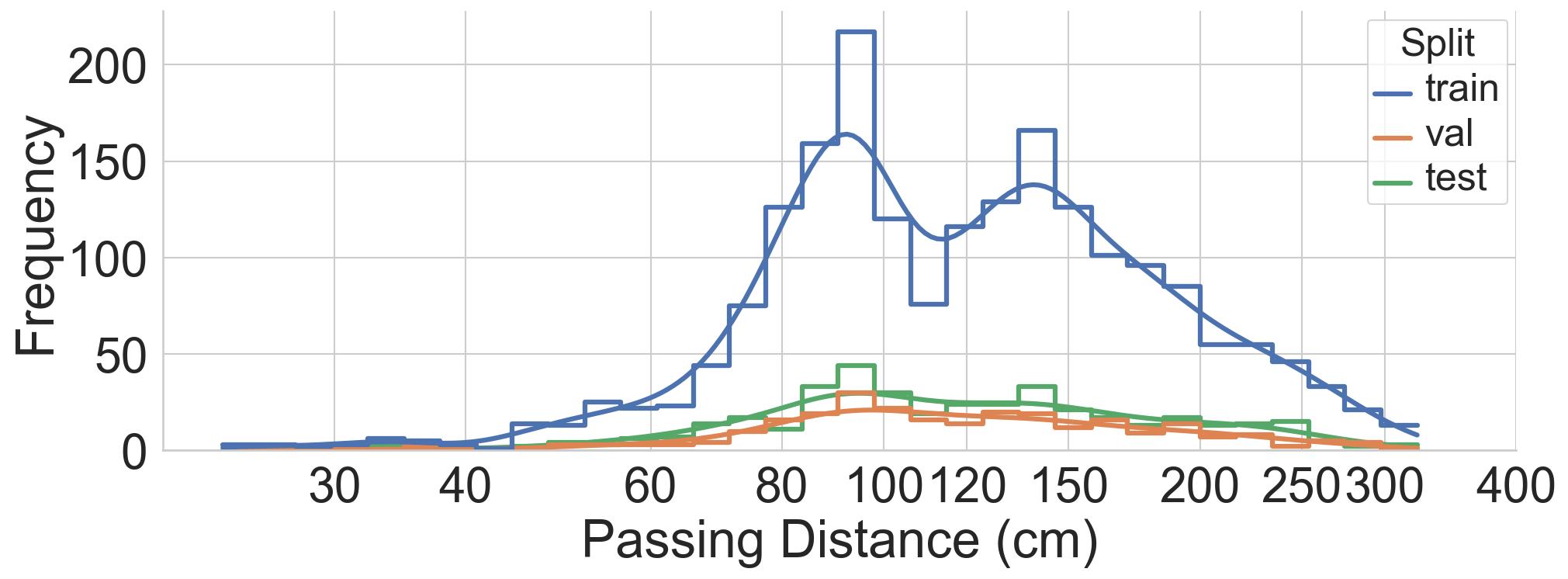}
\caption{Histogram of passing distances in our VOC training, validation, and testing subsets.}
\label{fig:voc_pdhist}
\end{figure}

In addition to the publicly available driving situation images and synthesized images, we also utilize a real-world on-road riding dataset, named Victorian On-road Cycling Dataset (VOC)~\cite{beck2019much}, to evaluate baseline models. The VOC dataset provides 330 hours of bicycle riding data obtained at various times of the day in various riding environments such as on-road and in shared bike and pedestrian paths. The data was collected from one forward-facing GoPro camera and rightward-facing ultrasonic distance sensors. The data collection rig was controlled through a Raspberry Pi 4 Single Board Computer (SBC) and we used its timestamp to synchronize the data streams. The front-view camera collected the whole cycling trip while distance sensors only recorded the lateral passing distance when an object was detected by the ultrasonic sensor in the capture time $t_c$ (see Figure.\ref{fig:criteria1}). All passing events were manually reviewed and additional details were annotated, including the category of passing vehicle and the speed limit zones of the road in which the passing event occurred. \mj{We segmented the original videos into clips based on the recorded distance measurement timestamps, ensuring that each clip contains a complete pass event. In total, we obtained 18,527 clips, and the mean passing distances by speed limit are presented in Figure.\ref{fig:VOC_boxplot}.} According to the local road rules, there were 1,342 illegal passing events, regarded as positive CP events in our task. For scene-level detection tasks, we randomly selected 1,300 legal passing events as negative samples for training and testing, more detailed label distributions in the VOC dataset can be found in Figure.\ref{fig:voc_countnm}. It is observed from Figure.\ref{fig:voc_pdhist} that most passing distances in our balanced VOC dataset are less than 1 meter which is illegal in any speed zone.
\section{Experiments}

\subsection{Experimental Details}

\begin{figure*}[th!]
    \centering
    \begin{subfigure}[b]{.32\textwidth}
        \includegraphics[width=\linewidth]{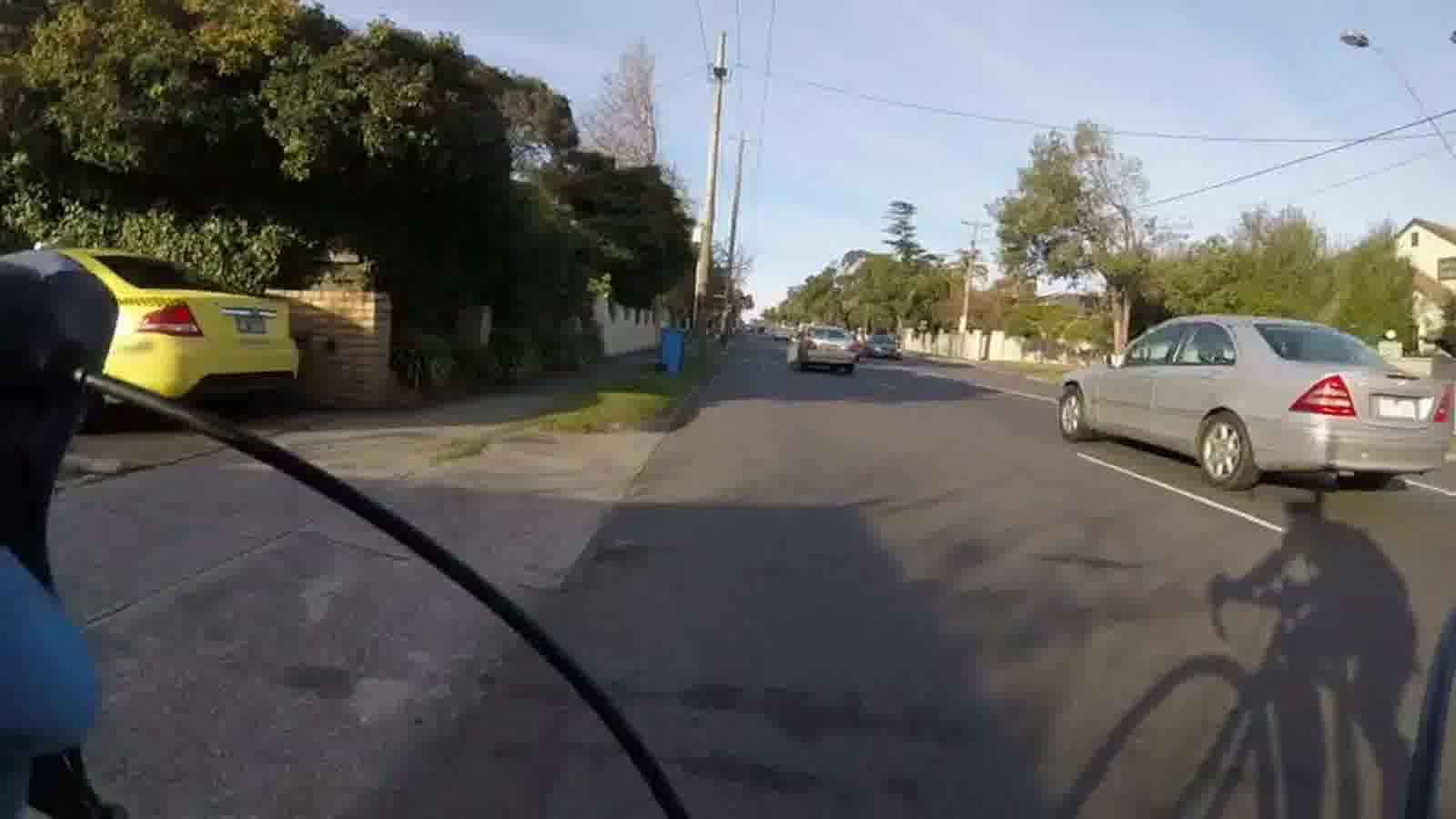}
        \caption{RGB Frame}
    \end{subfigure}
    \begin{subfigure}[b]{.32\textwidth}
        \includegraphics[width=\linewidth]{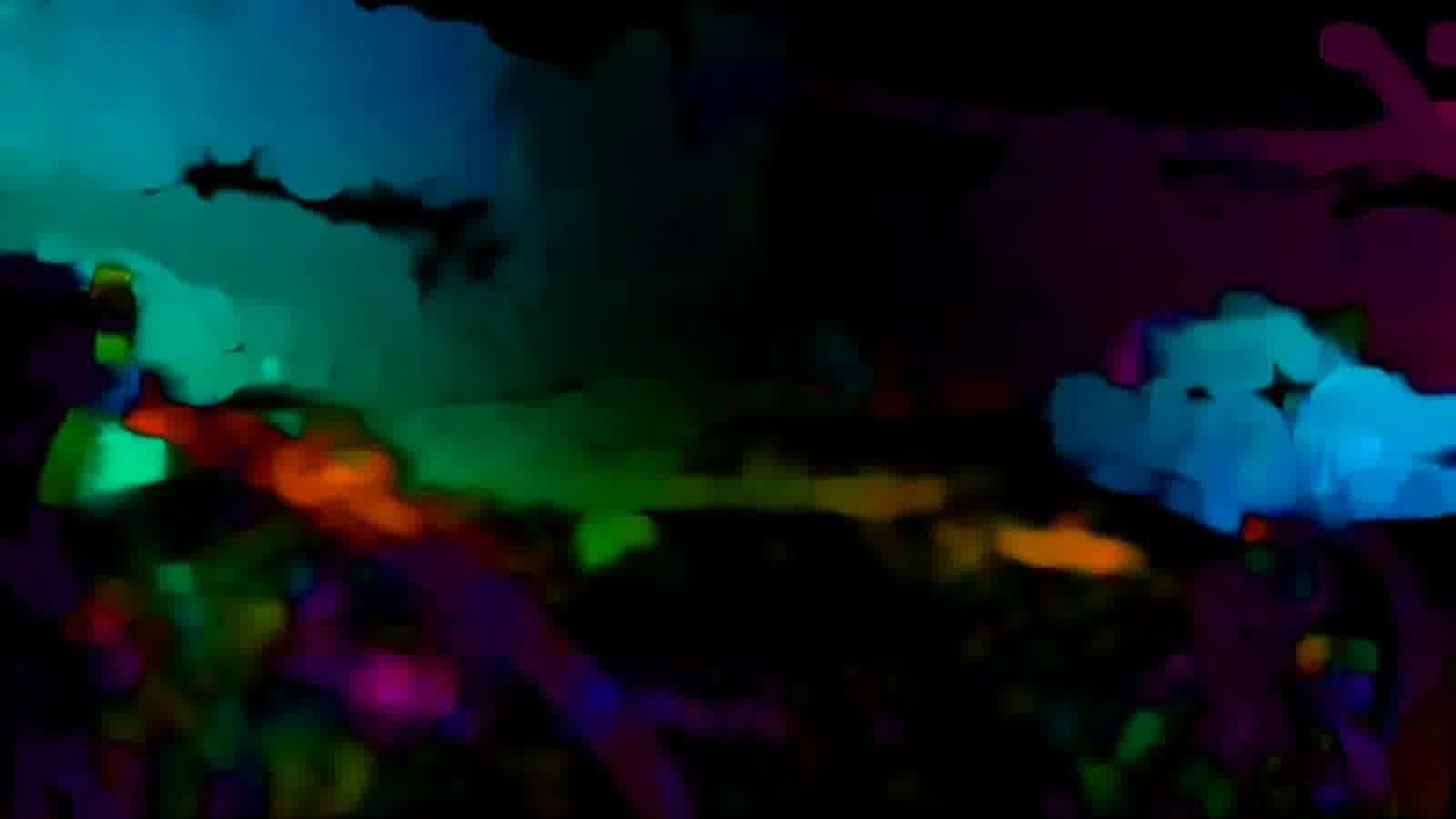}
        \caption{Optical Flow Frame}
        \label{subfig:input_opt}
    \end{subfigure}
    \begin{subfigure}[b]{.32\textwidth}
        \includegraphics[width=\linewidth]{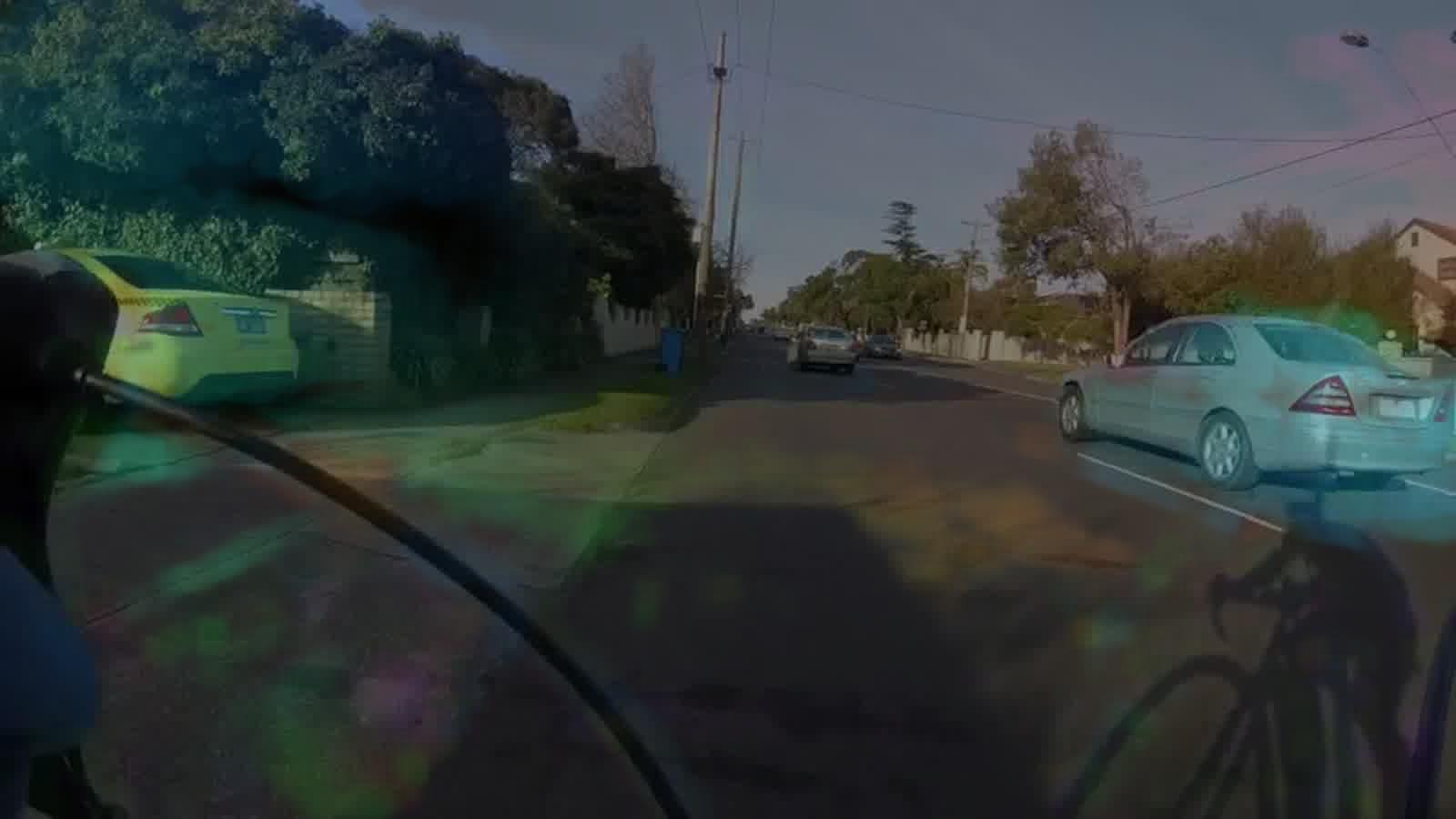}
        \caption{Fused Frame}
    \end{subfigure}
    \caption{Input Demonstration, where (a) is the RGB frame, (b) is the corresponding Franeback optical flow \cite{farneback2003two} which is either used as a separate input stream or used to calculate fused frame (c) following the method mentioned in \cite{ibrahim2021cyclingnet}.}
    \label{fig:input_types_demo}
\end{figure*}

All our models were implemented in PyTorch~\cite{pytorch} and trained on 2 GTX 2080Ti GPUs. We utilized FFmpeg~~\cite{ffmpeg} to extract 25 frames per second from videos in our VOC and CARLA synthetic datasets. For scene-level CP detection, only the VOC dataset is used and the input clip has 50 frames (the length is 2s which can cover a complete passing event). Frames were resized to $224\times224$ before being fed into models\footnote{We also performed hyper-parameter tuning, e.g., different frame sizes, clip lengths, and neural network structures, and found the results are robust to these hyper-parameter choices.}. In addition, to better understand the effect of using optical flow on the learning performance, we experiment with different input types as shown in Fig. \ref{fig:input_types_demo}, where single-stream feature extraction is used for taking either RGB or fused frame~\cite{ibrahim2021cyclingnet} as input, while two-streams feature extraction is used for taking both RGB and optical flow as input. Resnet101 pre-trained on ImageNet and I3D pre-trained on Kinetics are adopted as the backbone of two benchmark models, respectively. The LSTM in the CNN+LSTM model has two layers with 1024 hidden units and is bidirectional. Scene-level detection models are trained with an SGD optimizer with momentum set to 0.9 in an end-to-end manner. We set the learning rate to 0.001 and batch size to 16. All scene-level models are trained for 30 epochs where after each epoch the trained model is saved to a checkpoint. After the training, for each model, the checkpoint with the best performance on the validation dataset is utilized for inference. 
\begin{table*}[th!]
\centering
\caption{Best performances of scene-level CP detection models over 5 runs on our VOC dataset.}\label{tab:scene_performance}
\begin{tabularx}{\linewidth}{l|X|X|X|X|X|X}
\toprule
\multirow{2}{*}{Methods}  & \multicolumn{3}{c|}{Accuracy} & \multicolumn{3}{c}{F1-Score}     \\ \cline{2-7}
         & RGB & Fused & RGB \& OPT & RGB & Fused & RGB \& OPT \\ 
\hline
CNN+LSTM & \textbf{88.13} & 86.11 & 87.88 & \textbf{88.94} & 86.35 & 88.73  \\
I3D      & 84.60 & \textbf{85.10} & 82.32 & 84.71 & \textbf{85.29} &  82.05 \\
\bottomrule
\end{tabularx}
\end{table*}

\begin{figure}[th!]
    \centering
    \includegraphics[width=.9\linewidth]{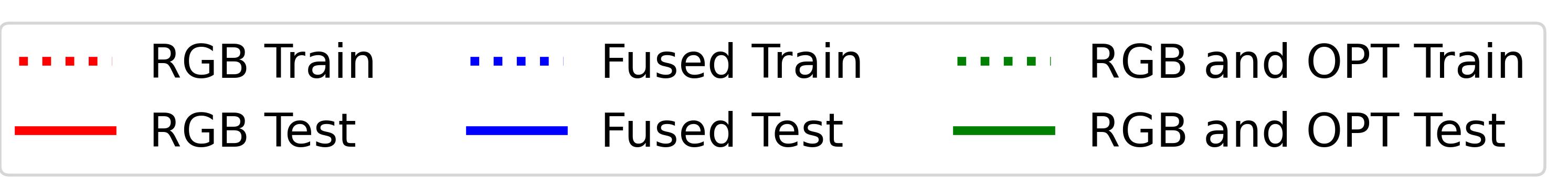}
    \includegraphics[width=\linewidth]{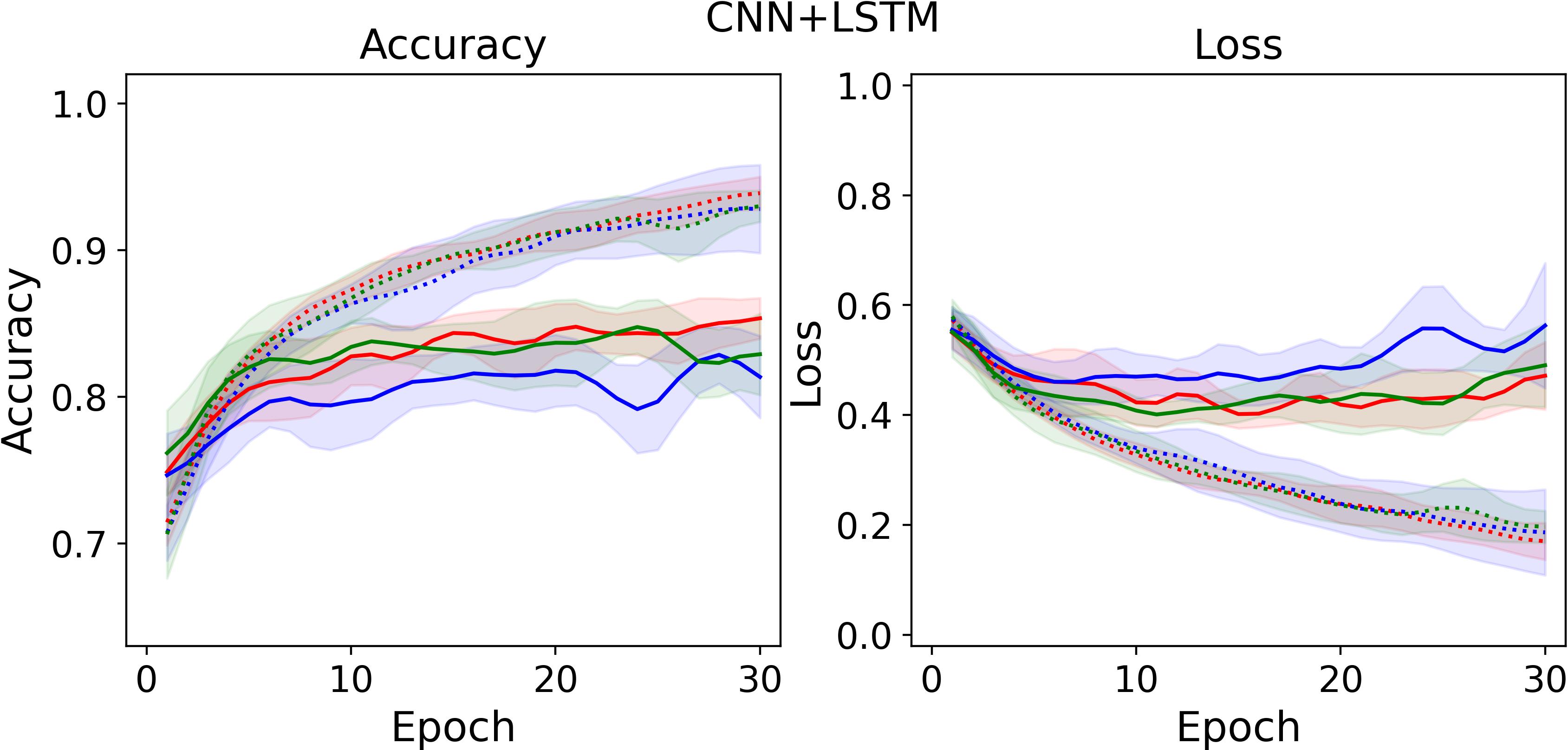}
    \includegraphics[width=\linewidth]{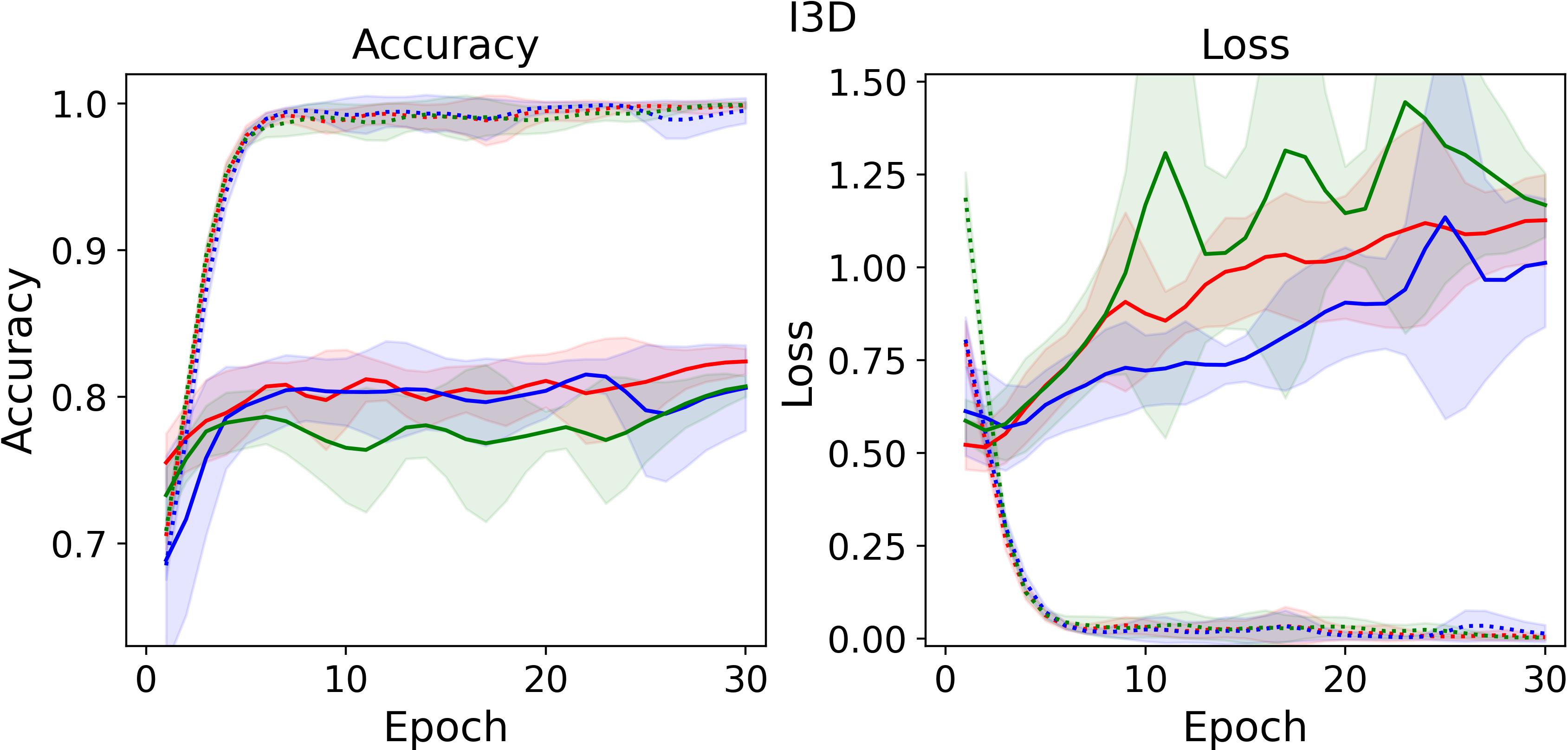}
    \caption{Performance Comparison On Various Input Types, where the lines correspond to the average over 5 runs and the shaded areas correspond to the areas within one standard deviation.}
    \label{fig:Performance_Comparison_On_Various_Input_Types}
\end{figure}
\begin{figure}[th!]
\centering
\includegraphics[width=.8\textwidth]{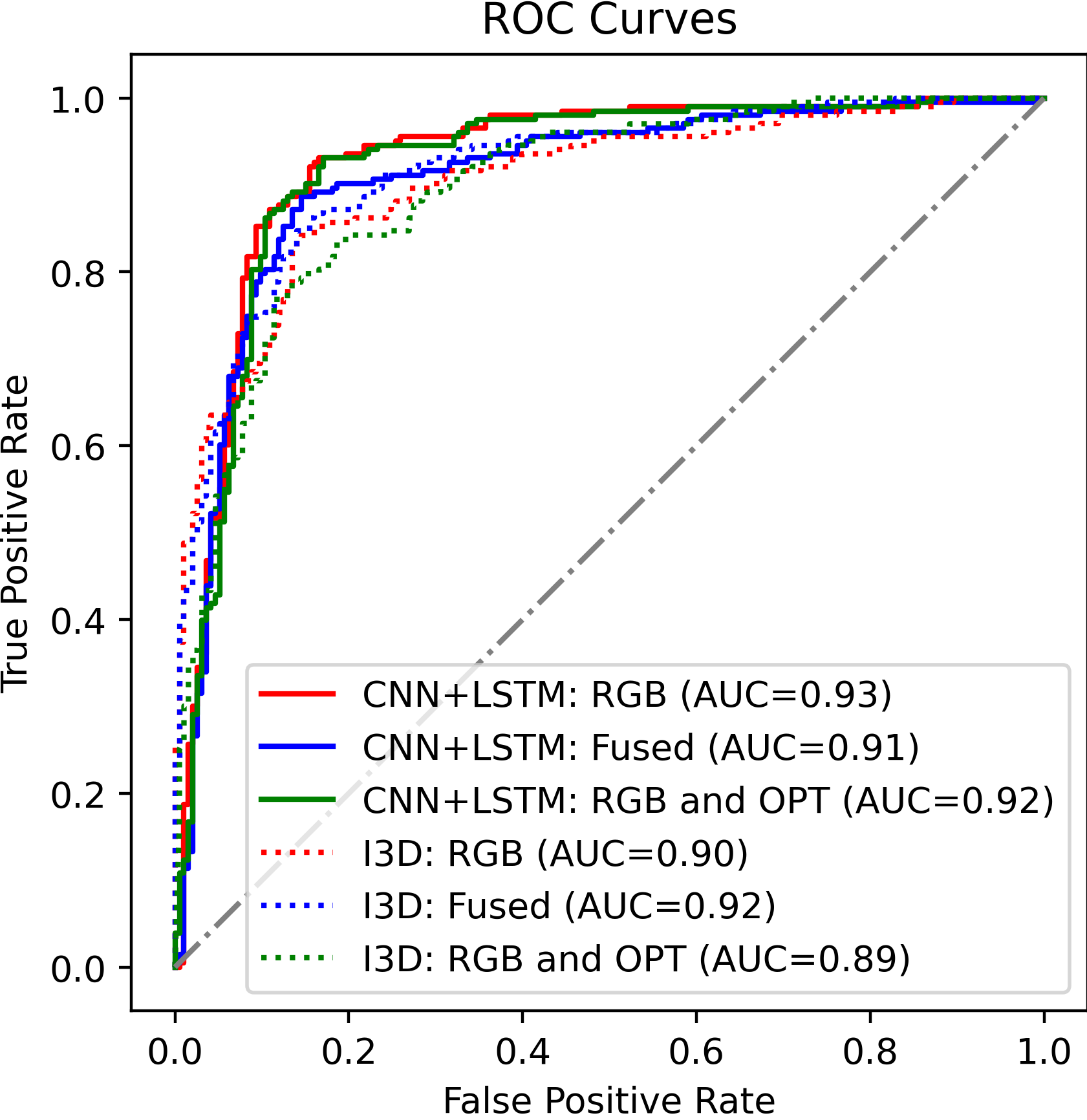}
\caption{Micro-average of receiver operating characteristic curve for scene-level CP detection from the real-world dataset.}
\label{fig:roc}
\end{figure}
For instance-level CP detection, the input frames have $1600\times900$ resolution. Our ICD is built on top of FCOS3D. We trained randomly initialized networks from scratch in an end-to-end manner. Models are trained with an SGD optimizer. Gradient clip and warm-up policy are exploited with a learning rate of 0.002, a warm-up of 500 steps, a warm-up ratio of 0.33, and batch size 16 on 2 GTX 2080Ti GPUs. We compare four kinds of training strategies: \textbf{NuScenes}, where ICD is trained on NuScenes dataset for 12 epochs; \textbf{Carla}, where ICD is trained on our Carla synthetic dataset for 20 epochs; \textbf{Finetune}, where ICD is firstly pre-trained on NuScenes for 12 epochs and then finetuned on Carla synthetic dataset for 4 epochs; \textbf{Alternating}, where ICD is trained on NuScenes for 3 epochs and Carla synthetic dataset for 1 epoch. We repeat this process for 4 times so that the model is trained for 16 epochs in total. Note that, we only finetune the regression branches and freeze the parameters in Resnet and FPN. In addition, because the VOC dataset does not have labeled bounding boxes, it is not used for training the instance-level model but is only used for the testing of the close passing event using the post-procedure introduced in \ref{subsubsec:postprocedure_ICD}.

\subsection{Scene-level CP Detection Results}

\subsubsection{Quantitative Analysis}


\begin{figure*}[th!]
\centering
\includegraphics[width=\textwidth]{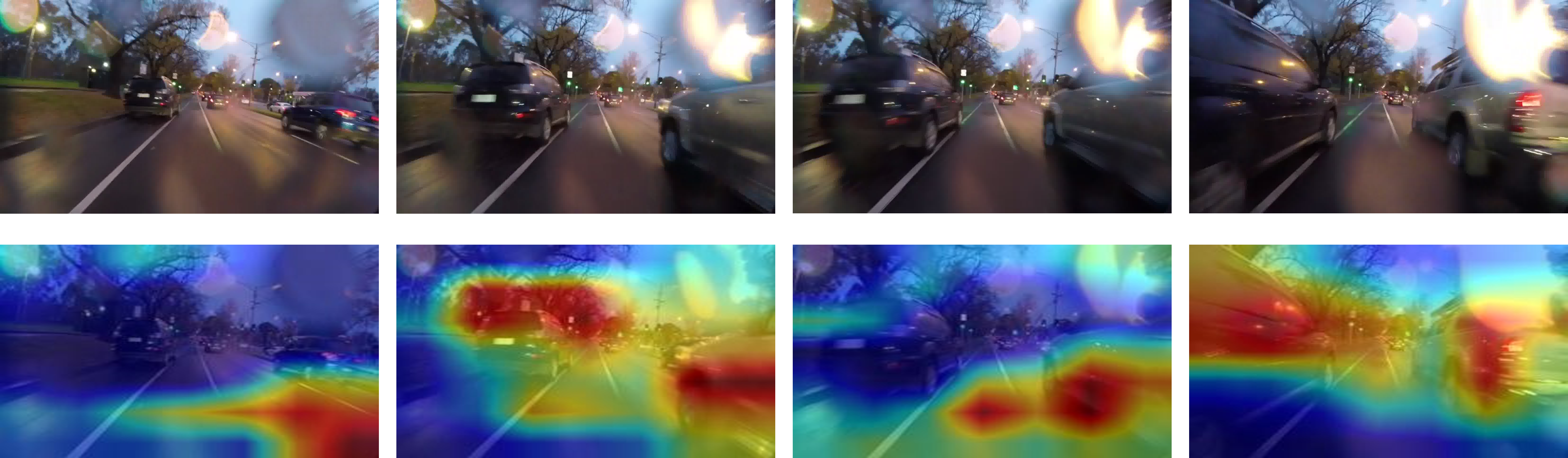}
\caption{Illustrations of scene-level detection based on our CNN+LSTM network. We also visualize CP activation maps in each frame.}
\label{fig:cnn_cam}
\end{figure*}

We evaluate the two scene-level detection models by standard classification metrics: \textit{i.e.} Accuracy, F1-score, and AUC. Fig. \ref{fig:Performance_Comparison_On_Various_Input_Types} and Table.\ref{tab:scene_performance} presents the training curves and the best performances of scene-level CPNM detection models on our VOC dataset. Figure.\ref{fig:roc} shows the micro-average of receiver operating characteristic curves (ROC) for the two models and presents AUC values. For both CNN+LSTM and I3D, adding optical flow as a separate input stream does not help in improving final performance, compared to only using RGB frames as input. I3D does have a slight performance improvement with fused input, while CNN+LSTM experiences a performance decrease with fused input. These findings indicate that using optical flow is not helpful for the two scene-level models. We speculate this is because of the less reliable optical flow caused by the unstable camera, as demonstrated in Fig. \ref{subfig:input_opt}. It is observed that the CNN+LSTM model achieved the benchmark performance on the scene-level detection task and outperformed the I3D model on all three metrics. The CNN+LSTM can achieve 88.13\% scene-level CP detection accuracy with RGB input on our VOC dataset which outperforms I3D with fused input by 3.03\%. As shown in Fig. \ref{fig:Performance_Comparison_On_Various_Input_Types} that the gap between the training and testing loss for I3D is much larger than that for CNN+LSTM, I3D is more susceptible to over-fitting than CNN+LSTM, even though they all employ similar dropout techniques to deal with the over-fitting problem. Different from CNN+LSTM where ResNet is employed to extract features and fixed during the training, I3D is trained from scratch for its feature extraction, which makes it susceptible to over-fitting. In addition, the differences in the structures of these two models also contribute to the performance difference. For instance, the CNN+LSTM network has the temporal information processing capabilities to recognize a passing event due to the LSTM layer. More importantly, the 2D CNN in CNN+LSTM extracts spatial features from each frame and endows the whole model with capabilities to measure the distance between a passing vehicle and camera, while the 3D CNN in I3D tries to simultaneously extract both spatial and temporal information which may confuse when allocating gradients. Those differences make CNN+LSTM superior to I3D.

\subsubsection{Qualitative Analysis}

Figure.\ref{fig:cnn_cam} illustrates a positive CP sample from our VOC dataset and the prediction from the CNN+LSTM network. Firstly, the displayed frames reveal that the VOC dataset provides challenging situation videos, like in rainy weather and dark environments. Training on challenging videos improves the model's robustness, the network can give accurate predictions when input frames are noisy (where the rain blurs objects). The second row in Figure.\ref{fig:cnn_cam} shows activation maps when the CNN+LSTM network made a prediction, generated by grad-cam~\cite{selvaraju2017grad} to increase the explainability of the detection model. We argue that the model should focus on the right side of each frame since right-hand drive vehicles overtake the bike from the right side. The first picture in the second row illustrates this argument. When there are no passing vehicles, the model will pay attention to the right bottom region where a vehicle may appear. The last three pictures show that the CNN+LSTM network focuses on the passing vehicle when it appears. 

\begin{table}[th!]
\centering
\caption{Accuracy of our proposed ICD under different training strategies on our VOC dataset.}\label{tab:icd_per}
\begin{tabular}{l|llll} 
\toprule
Strategy & NuScenes & Carla & Finetune & Alternating  \\
\hline
Accuracy & 35.60    & 8.58  & 77.28    & 84.09        \\
\bottomrule
\end{tabular}
\end{table}

\begin{figure*}[th!]
\centering
\includegraphics[width=\textwidth]{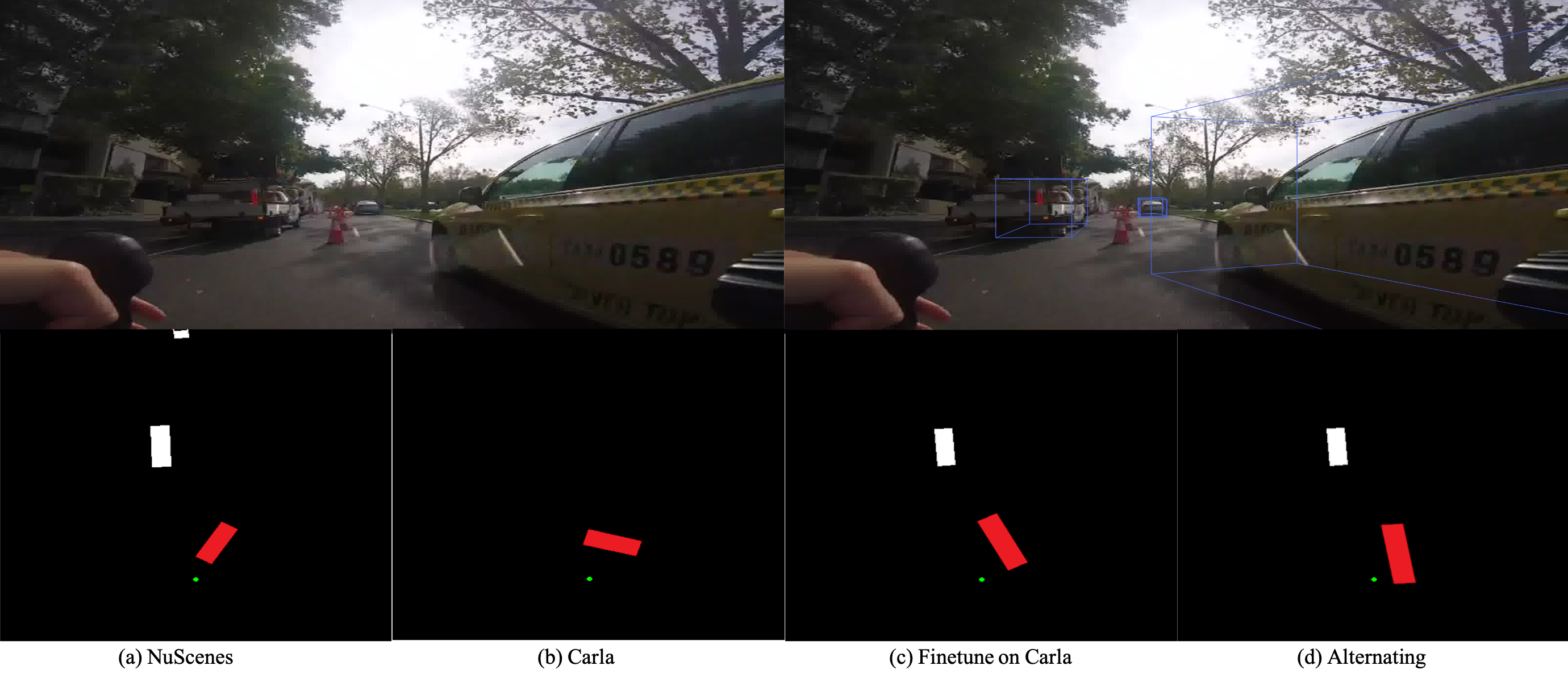}
\caption{Illustrations of instance-level detection based on our ICD model. We also visualize the bird eye view of ICD trained with different strategies. The green circle represents our camera, red and white rectangles refer to the CP and legal passing vehicles.}
\label{fig:brv}
\end{figure*}

\subsubsection{\mj{Large Multi-modal Model Performances}}

\begin{table}
\centering
\caption{\mj{Performance of InternVideo 2.5 in zero-shot and few-shot scene-level CP Detection}}\label{tab:LLM}
\begin{tabular}{l|ccccc} 
\hline
Setting  & Zero-shot & 1-shot & 5-shot & 10-shot & 20-shot  \\ 
\hline
Accuracy & 62.9\%    & 68.3\% & 70.5\% & 71.3\%  & 71.7\%   \\
\hline
\end{tabular}
\end{table}

\mj{Table~\ref{tab:LLM} presents the performance of InternVideo 2.5 in zero-shot and few-shot settings for scene-level CP detection. The results indicate that zero-shot accuracy is 62.9\%, suggesting that without any prior examples, the model can still recognize some CP events based on its pre-trained knowledge and general video understanding. As the number of shots increases, accuracy improves steadily, reaching 68.3\% in 1-shot, 70.5\% in 5-shot, and peaking at 71.7\% in 20-shot. However, we observe that after 10-shot learning, the performance gain becomes marginal, suggesting that additional context beyond this threshold does not significantly improve CP detection accuracy.}

\mj{Compared to trainable CNN-based video recognition models, InternVideo 2.5 underperforms in overall accuracy, as traditional models trained specifically on the CP dataset can better capture event-specific features. One possible reason is that VLMs rely on general video understanding rather than direct training on CP events, making them less optimized for this specific task. Additionally, VLMs may struggle with subtle CP definitions and variations due to their dependence on natural language prompts rather than direct feature learning, as well as lack of geometric spatial grounding. Despite these limitations, InternVideo 2.5 offers unique advantages such as interactivity and interpretability, which traditional CNN-based models lack. As VLMs continue to evolve, their video understanding capabilities are expected to improve, making them a promising direction for future CP detection tasks.}

\subsection{Instance-level CP Detection Results}

\subsubsection{Quantitative Analysis}

\begin{figure}[th!]
\centering
\includegraphics[width=\textwidth]{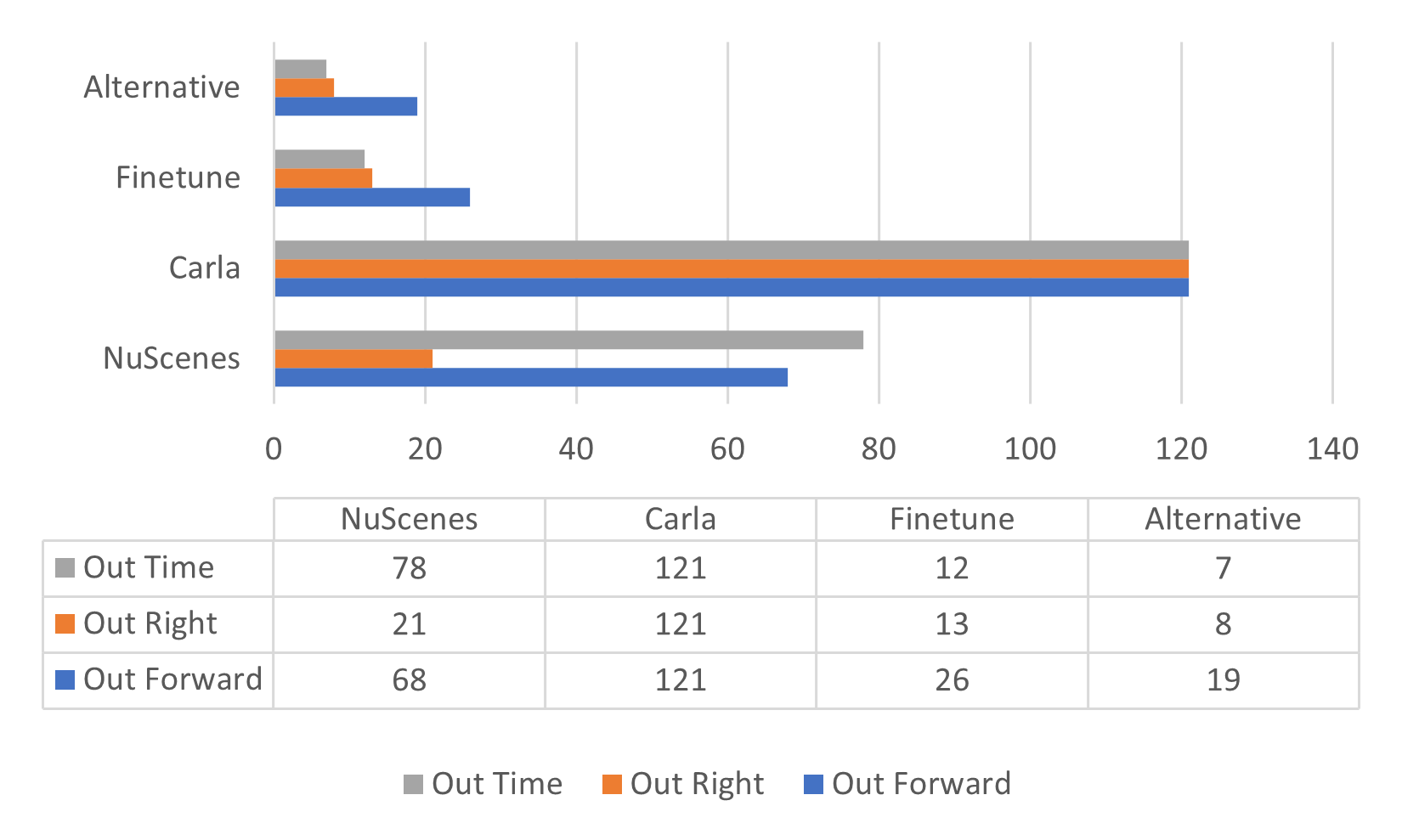}
\caption{Error types analysis of ICD with 4 different training strategies for instance-level CP detection on our VOC dataset.}
\label{fig:error_type}
\end{figure}

Table.\ref{tab:icd_per} summarizes the prediction accuracy of our proposed ICD on the VOC dataset with different training strategies, where VOC is not used during training but only for testing because of its lack of labeled bounding boxes. We observe that ICD with the Alternating training strategy achieves the highest accuracy. It is also clear that both Alternating and Finetune strategies significantly outperform the other two training strategies. By comparing the results of the first and third columns, fine-tuning the ICD can boost the performance of NuScenes training by 41.68\%. The alternating learning strategy can further increase the accuracy to 84.09\%. ICD is built on a monocular 3D object detection network which is extremely data-greedy. NuScenes can provide a large amount of real-world traffic situation data. Training ICD on NuScenes primes the model's capability to recognize real-world vehicles. However, regression branches that are used to transform 2D features to 3D information are trained in driving scenes. When transferring such a model to the VOC dataset, it fails to predict accurate 3D locations and sizes. In contrast, Carla's synthetic dataset can provide cycling scenes, which facilitates learning the 3D transformations. However, the vehicles in Carla look different from real-world vehicles. Therefore, a model trained solely on the Carla synthetic dataset is unable to recognize real-world objects, and this is the reason it performs the worst on the VOC dataset. 

We proposed three criteria for instance-level CP detection, to further compare the 4 training strategies, we conducted an error type analysis and presented the results in Figure.\ref{fig:error_type}. \textit{Out Right} refers to the detected objects that fail to fulfill the first criterion. \textit{Out Forward} refers to the objects that break the second criterion. \textit{Out Time} refers to the objects that fulfill all the criteria but are out of the passing event time range. For example, if a passing event is recorded at time $t_c$, the time $t$ we detect a CP should be in the following range:
\begin{align}
    t_c-0.4 < t < t_c+1.2
\end{align}
It is observed that our ICD can accurately predict the right-forward values, like the right distance and the width, which is better than predicting forward values, like the length and forward distance. We speculate that different camera parameters affect the transformation performances. We believe this can be alleviated in future work by either adding more synthetic data to the CARLA or creating a real-world on-road cycling dataset with 3D annotations, or ideally using a combination of real and synthetic data to avoid the need for extensive annotations.

\subsubsection{Qualitative Analysis}

To further investigate ICD with different training strategies qualitatively, we randomly select a CP sample from our VOC test set and visualize the bird eye view (BEV). The ICD algorithm first detects vehicles from the given input frame and then predicts their 3D sizes and locations. We visualize those 3D sizes and locations in the BEV, as shown in Fig. \ref{fig:brv}. ICD trained on NuScenes can detect the vehicles but predicts incorrect allocentric orientation angles. We hypothesize this is due to the differences in the camera configurations between the two datasets. Training on Carla only fails to detect objects and predict their 3D sizes and locations. The (c) and (d) pictures reveal that training the feature extraction modules and regression branches on NuScenes and Carla synthetic datasets respectively can significantly improve performance. Then the proposed criteria are used to identify whether a vehicle is passing illegally or legally. The vehicles identified as making a close pass are colored in red. In the right top image, we localize the detected vehicles with 3D bounding boxes. This observation suggests that ICD is effective at detecting objects from riding scenes.

\section{Discussion}


Detecting CP events presents considerable challenges due to several factors: First, the quality of video data can significantly fluctuate based on varying lighting and weather conditions. In extreme scenarios, these conditions can severely hinder the detection of vehicles within the scene, thereby compromising the efficacy of CP detection algorithms. Furthermore, the constrained viewing angles of forward-facing cameras, which are commonly employed for data collection, pose additional difficulties in capturing comprehensive scenes. The vehicle's passing distance cannot be accurately assessed when it lies outside the camera's field of view; detection only becomes feasible once the vehicle overtakes the cyclist and enters the camera's visual range. Moreover, camera motion exacerbates these challenges. Vibrations transmitted through the bicycle's handlebars to the mounted camera can cause the resulting images to blur or videos to exhibit pronounced wobble, further detracting from the quality of the data and, by extension, the performance of detection algorithms. These factors collectively underscore the complexities involved in CP event detection and highlight the need for robust methodologies capable of mitigating these impediments to ensure accurate and reliable identification of CP events. 

\mj{Alternative sensors could be considered for capturing precise proximity data. For example, LIDAR sensors mounted at both the front and rear of the bike could enable 360-degree depth data capture around the cyclist, rather than relying on the single-point estimate obtained from ultrasound. This approach could provide a more comprehensive spatial understanding of vehicle-cyclist interactions and improve the reliability of close-pass measurements. However, the size, mass and power requirements of additional sensing may impact the cyclist's behavior as well as the other road users.}

Another significant challenge is the collection of reliable labels for CP events. The conventional methodology involves the engagement of human experts to undertake the manual annotation of video data. This process, while beneficial for ensuring the accuracy of labels, is inherently time-intensive and susceptible to subjective biases, potentially affecting the consistency and objectivity of the annotations provided~\cite{vansteenkiste2016hazard}. Furthermore, the dataset may be unbalanced, with a much greater number of negative samples (legal events) compared to positive samples (CP events). Those unbalances can lead to biased models that prioritize negative samples and result in low sensitivity to CP events.

To deal with these challenges, this study introduced two additional benchmark datasets to facilitate the advancement of CP event detection, including a real-world on-biking dataset with CP events labeled, \textit{i.e.}, VOC, and a synthetic cycling dataset, i.e., CARLA. Furthermore, two types of benchmark models are investigated in this work, considering CP detection accuracy at the scene level and increasing the interpretability of the detection at the instance level. Particularly, because instance-level detection requires more fine-grained labelling compared to the scene-level detection, in this work we propose a hybrid training approach where the combination of the autonomous driving dataset and the synthetic dataset is used to train the instance-level model to reduce the reliance on costly 3D annotations. Initial findings reveal that the prediction accuracy of the instance-level model on the real-world on-biking dataset marginally trails that of the scene-level models, while not requiring any real-world data during training. This observation underscores the potential for further refinement of the instance-level model's predictive accuracy through the integration of additional synthetic and real-world on-biking data. Such enhancements are anticipated to not only improve the model's performance but also contribute significantly to the development of more sophisticated and reliable CP event detection systems. \mj{Additionally, we have observed that utilizing LMMs for video reasoning can help us preliminarily filter other hazardous interactions from large volumes of cycling videos when given appropriate prompts. This capability could significantly aid in identifying diverse road hazards beyond close passes. However, due to the high computational cost and time required for processing extensive video datasets with LMMs, we recognize the need to explore more efficient algorithms to scale this approach effectively. This is an important avenue for future research, and we will continue investigating ways to optimize computational efficiency while maintaining detection accuracy.}

A limitation of the proposed benchmark is that it focuses only on CP events\mj{, which is one of the most frequent near-miss events on the road~\cite{aldred2016cycling}.}. While expanding to other types of near-miss events would be valuable, the definition of other near-miss events can be ambiguous and subjective. This is especially challenging given the limited sensor and perception capacity, which makes it difficult to detect all possible factors in the scene. Furthermore, the study did not investigate whether data augmentation techniques could be used to deal with the challenge of unbalanced datasets.

\section{Conclusion and Future Work}
In this paper, we propose a novel benchmark, named Cyc-CP, for CP detection from video streams. In this benchmark, CP is defined as an illegal close pass according to local road rules, objectively. Our Cyc-CP first proposes two formulations for CP detection, namely scene-level and instance-level detection. For each task, we propose baseline and benchmark models relying on deep learning techniques. Those models are trained and evaluated on the public NuScenes dataset, Victorian On-road Cycling dataset, and CARLA synthetic dataset. We hope that our Cyc-CP can be used to improve the development of CP detection systems to help cyclists reduce the risks on the road and assist policy-makers in better understanding cycling safety measurements.

\mj{Future research should expand beyond CP events to cover a broader range of near-miss incidents, incorporating various sensing modalities to improve real-world data collection. In addition to using simulation tools to generate diverse training samples, different sensors and camera setups such as LiDAR should be investigated to enhance real-world data acquisition. This will enable us to capture more detailed spatial and contextual information, improving the robustness of our dataset. Moreover, we plan to enhance our baseline models by integrating object-level features into scene-level detection, allowing for a deeper understanding of the interactions between road users. We will also investigate data augmentation techniques to address class imbalances and increase dataset diversity. Through these advancements, we aim to develop a more comprehensive and reliable near-miss detection system, capable of effectively analyzing hazardous interactions in real-world cycling environments.}


\section{Acknowledgement}
This project was funded by an Australian Government Department of Infrastructure, Transport, Regional Development and Communications `Road Safety Innovation Fund' grant. Ben Beck was supported by an Australian Research Council Future Fellowship (FT210100183). Dana Kuli\'c was supported by an Australian Research Council Future Fellowship (FT200100761). Xiaojun Chang was supported by the Australian Research Council (ARC) Discovery Early Career Researcher Award (DECRA) DE190100626. This research was enabled in part by support provided by Compute Canada (www.computecanada.ca).

\bibliographystyle{IEEEtran}
\bibliography{ref}

\end{document}